\documentclass[journal]{IEEEtran}

\usepackage[pagebackref=true,breaklinks=true,letterpaper=true,colorlinks,bookmarks=false]{hyperref}

\usepackage{times}
\usepackage{epsfig}
\usepackage{graphicx}
\usepackage{amsmath}
\usepackage{amssymb}
\usepackage{multirow}
\usepackage{amsthm}
\usepackage{subfigure}
\usepackage{color}
\usepackage{flushend}

\newcommand\eg{\emph{e.g.}} 
\newcommand\ie{\emph{i.e.}}

\newcommand\etal{\emph{et al.}}

\begin{document}

\title{Deep Edge-Aware Saliency Detection}

\author{Jing~Zhang,~\IEEEmembership{Student Member,~IEEE,}
Yuchao~Dai,~\IEEEmembership{Member,~IEEE,}~Fatih~Porikli,~\IEEEmembership{Fellow,~IEEE} and~Mingyi~He,~\IEEEmembership{Senior Member,~IEEE}
\thanks{Jing Zhang and Mingyi He are with School of Electronics and Information, Northwestern Polytechnical University, China. Jing Zhang is currently visiting the Australian National University supported by the China Scholarship Council.}
\thanks{Yuchao Dai and Fatih Porikli are with Research School of Engineering, Australian National University, Australia.}
\thanks{This work was supported in part by Natural Science Foundation of China grants (61420106007, 61671387) and the Australian Research Council (ARC) grants (DE140100180, DP150104645).%
}
}


\maketitle

\begin{abstract}
There has been profound progress in visual saliency thanks to the deep learning architectures, however, there still exist three major challenges that hinder the detection performance for scenes with complex compositions, multiple salient objects, and salient objects of diverse scales. In particular, output maps of the existing methods remain low in spatial resolution causing blurred edges due to the stride and pooling operations, networks often neglect descriptive statistical and handcrafted priors that have potential to complement saliency detection results, and deep features at different layers stay mainly desolate waiting to be effectively fused to handle multi-scale salient objects. In this paper, we tackle these issues by a new fully convolutional neural network that jointly learns salient edges and saliency labels in an end-to-end fashion. Our framework first employs convolutional layers that reformulate the detection task as a dense labeling problem, then integrates handcrafted saliency features in a hierarchical manner into lower and higher levels of the deep network to leverage available information for multi-scale response, and finally refines the saliency map through dilated convolutions by imposing context. In this way, the salient edge priors are efficiently incorporated and the output resolution is significantly improved while keeping the memory requirements low, leading to cleaner and sharper object boundaries. Extensive experimental analyses on ten benchmarks demonstrate that our framework achieves consistently superior performance and attains robustness for complex scenes in comparison to the very recent state-of-the-art approaches.

\end{abstract}


\IEEEpeerreviewmaketitle

\section{Introduction}

\begin{figure}[!t]
   \begin{center}
   \begin{tabular}{ c@{ } c@{ } c@{ } c@{ }  c@{ }  c@{ } c@{ }}
   {\includegraphics[width=0.125\linewidth]{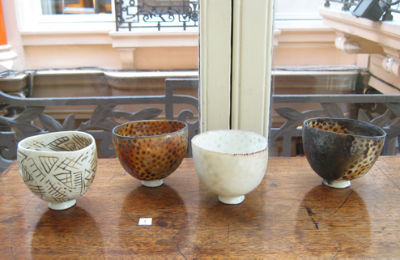}} &
   {\includegraphics[width=0.125\linewidth]{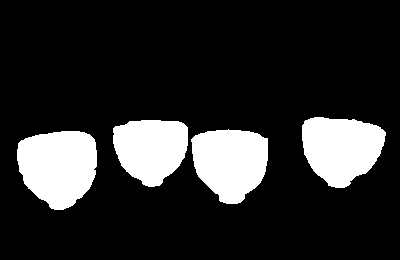}}&
   {\includegraphics[width=0.125\linewidth]{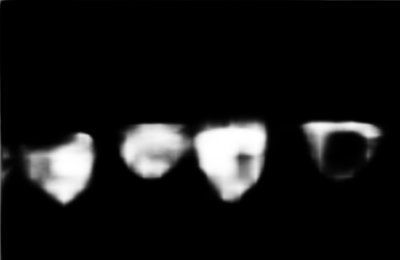}}&
   {\includegraphics[width=0.125\linewidth]{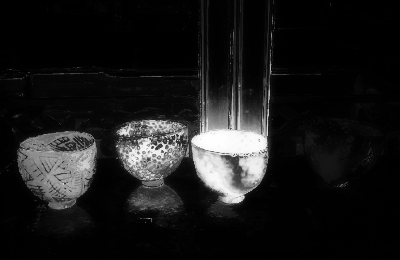}} &
   {\includegraphics[width=0.125\linewidth]{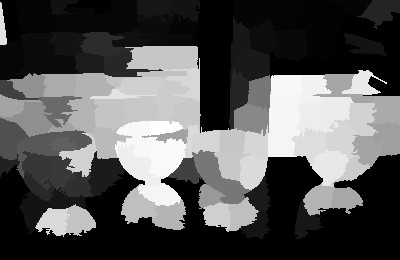}}&
   {\includegraphics[width=0.125\linewidth]{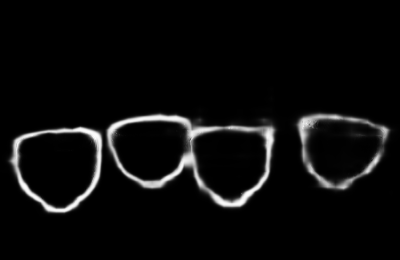}}&
   {\includegraphics[width=0.125\linewidth]{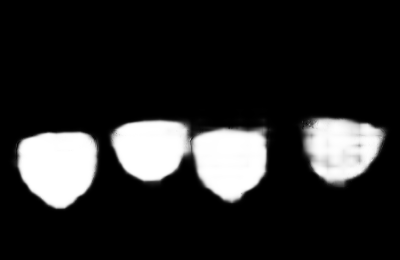}}　\\
      {\includegraphics[width=0.125\linewidth]{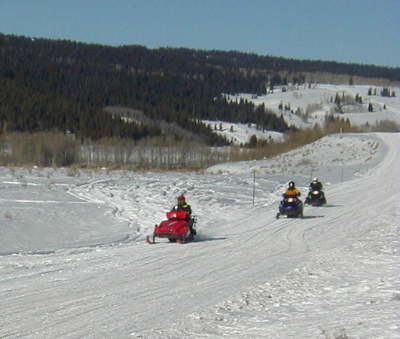}}&
   {\includegraphics[width=0.125\linewidth]{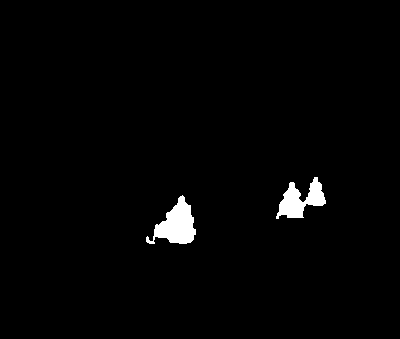}}&
   {\includegraphics[width=0.125\linewidth]{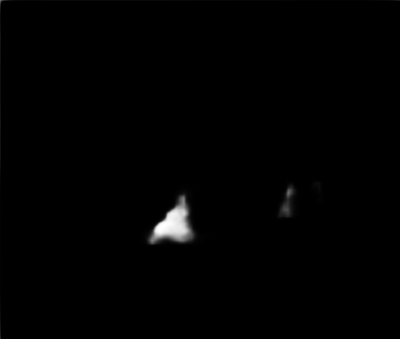}}&
   {\includegraphics[width=0.125\linewidth]{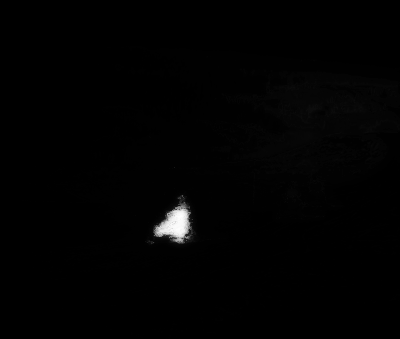}}&
   {\includegraphics[width=0.125\linewidth]{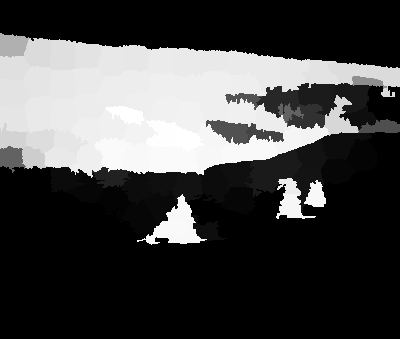}}&
   {\includegraphics[width=0.125\linewidth]{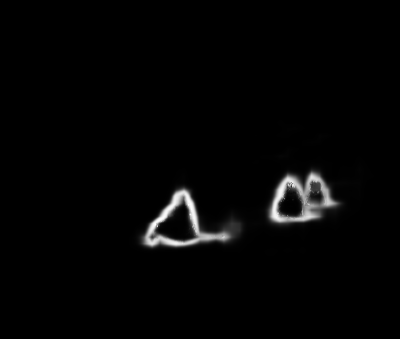}}&
   {\includegraphics[width=0.125\linewidth]{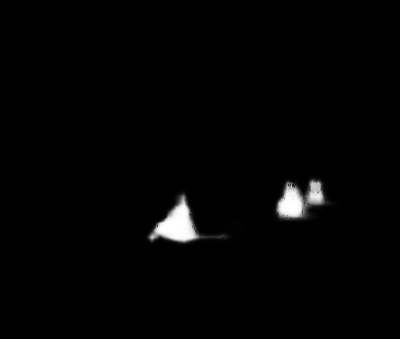}} \\
      {\includegraphics[width=0.125\linewidth]{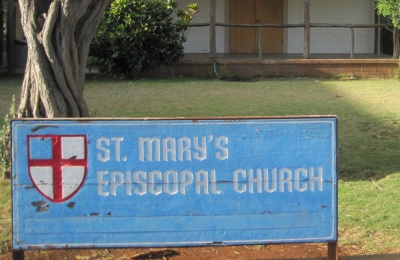}}&
   {\includegraphics[width=0.125\linewidth]{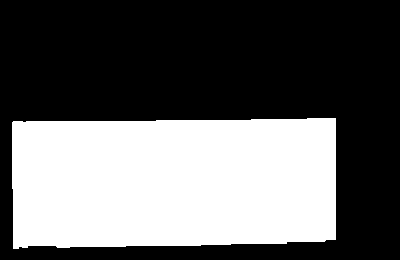}}&
   {\includegraphics[width=0.125\linewidth]{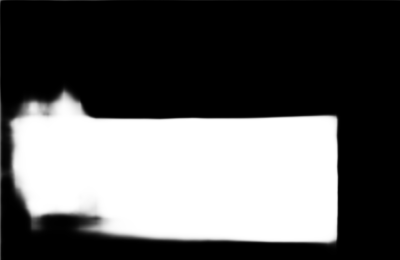}}&
   {\includegraphics[width=0.125\linewidth]{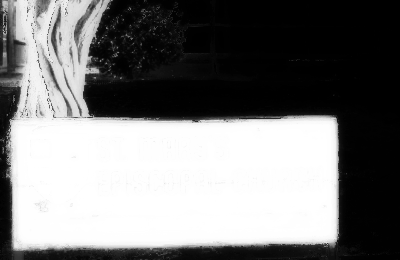}}&
   {\includegraphics[width=0.125\linewidth]{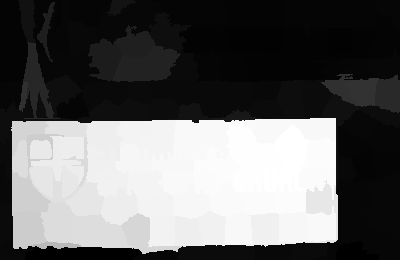}}&
   {\includegraphics[width=0.125\linewidth]{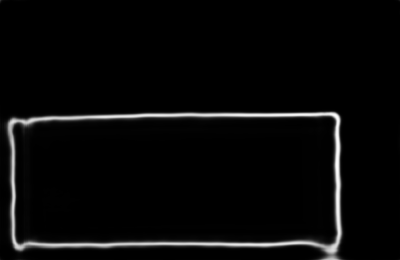}}&
   {\includegraphics[width=0.125\linewidth]{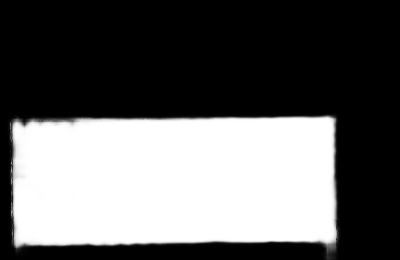}} \\
   \footnotesize{Image} & \footnotesize{GT} & \footnotesize{DSS} & \footnotesize{DC} & \footnotesize{RBD} & \footnotesize{Edge} & \footnotesize{Ours}\\
   \end{tabular}
   \end{center}
   \vspace{-2mm}
\caption{Various challenging complex real world scenarios for saliency detection, which either depict multiple salient objects (top row), or salient objects with diverse scales (middle row and bottom row). From left to right: Input image, ground truth, result of DSS \cite{ChengCVPR17}, DC \cite{DC}, RBD \cite{Background-Detection:CVPR-2014}, salient edge map and saliency map of our method.}
   \label{fig:saliency_compare1}
\end{figure}

\IEEEPARstart{S}{aliency} detection (salient object detection) \cite{Dataset-Baseline:TIP-2015}, \cite{MC_ICCV13}, \cite{Soft-Image-Abstraction:ICCV-2013} aims at identifying the visually interesting object regions that are consistent with human perception. It is intrinsic to many computer vision tasks such as image cropping \cite{image_cropping_2009}, context-aware image editing \cite{image_editing_2009}, image recognition \cite{image_recognition_2006}, interactive image segmentation \cite{Segmentation_saliency}, action recognition \cite{sharma2015attention}, image caption generation \cite{Xu2015show} and semantic image labeling \cite{SemanticLabel}. Albeit considerable progress, it still remains as a challenging task and requires effective approaches to handle complex real-world scenarios (see Fig.~\ref{fig:saliency_compare1} for various examples).

Conventional saliency detection methods either employ predefined features such as color and texture descriptors \cite{DRFI:CVPR-2013}\cite{YehPR}, or indicators of appearance uniqueness \cite{UFO-Saliency:ICCV-2013} and region compactness \cite{Soft-Image-Abstraction:ICCV-2013} based on specific statistical priors such as center prior \cite{ContextSaliency}, contrast prior \cite{High-Dim-Color-Transform:CVPR-2014}, boundary prior \cite{Background-Detection:CVPR-2014} and object prior \cite{Huo2016162}. These handcrafted methods achieve acceptable results on relatively simple datasets (see \cite{Salient-Detection-Survey:2014} for a dedicated survey on saliency detection prior to the deep learning revolution), but their performances deteriorate quickly when the input images become cluttered and complicated.

Data-driven approaches, in particular, deep learning with convolutional neural networks (CNNs), have recently attained great success in many computer vision tasks such as image classification \cite{ResHe2015} and semantic segmentation \cite{Deeplab}. They have been naturally extended to saliency detection, where the problem is often formulated as a dense labeling task that automatically learns feature representations of salient regions, outperforming handcrafted solutions with a wide margin \cite{ChengCVPR17}, \cite{TIP}, \cite{MDF:CVPR-2015}, \cite{DeepMC}, \cite{DC}, \cite{RFCN}, \cite{DISC}, \cite{RACDNN}.


Albeit profound progress thanks to deep learning architectures, there still exist three major challenges that hinder the performance of deep saliency methods under complex real-world scenarios, especially for scenes depicting multiple salient objects and salient objects with diverse scales:
\begin{itemize}

\item \textbf{Low-resolution output maps:} Due to stride operation and pooling layers in CNN architectures, the resultant saliency maps are inordinately low in spatial resolution, causing blurred edges as illustrated in Fig.~\ref{fig:saliency_compare1}. The resolution of saliency maps is critical to several vision tasks such as image editing \cite{image_editing_2009} and image segmentation \cite{Segmentation_saliency}.

\item \textbf{Missing handcrafted yet pivotal features:} Deep learning networks neglect the statistical priors widely used in handcrafted saliency methods. Such features are often based on human intuition and applicable a wide-spectrum of cases. In Fig.~\ref{fig:saliency_compare1}, we provide examples where handcrafted saliency method outperforms the deep saliency counterparts. This reveals the complementary relation of human-knowledge driven and data-driven schemes and motivates us to seek better fusion approaches.

\item \textbf{Archaic handling of multi-scale saliency:} The implementations of existing deep learning based saliency methods do not effectively exploit features at different levels. Bottom-level features (information about details) are usually underestimated and top-level features (semantic cues) are overestimated. In the middle row of Fig.~\ref{fig:saliency_compare1}, we show examples where deep learning methods fail to capture the whole salient regions when the salient objects are very small.
\end{itemize}

\begin{figure*}[!htp]
   \begin{center}
   {\includegraphics[width=0.9\linewidth]{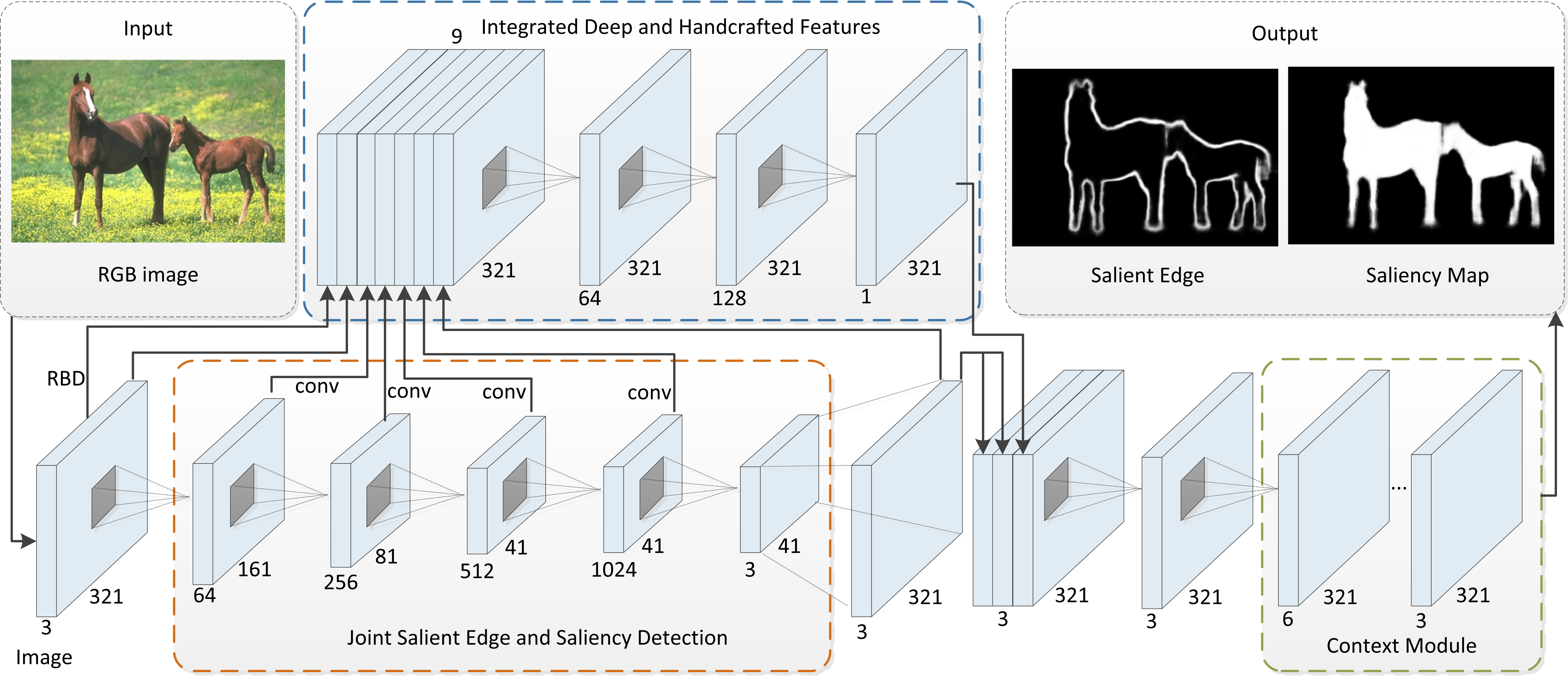}}
   \end{center}
   \vspace{-2mm}
   \caption{In a nutshell, our framework consists of three modules: 1) a front-end deep FCN module to jointly learn salient edge and saliency map; 2) a feature integration module to fuse handcrafted saliency features and deep saliency features across different scales; and 3) a context module to refine the saliency map.}
   \label{fig:Nutshell_Overview}
\end{figure*}

To tackle the above challenges, we propose a fully convolutional neural network (FCN) and an end-to-end learning framework for edge-aware saliency detection, as depicted in Fig.~\ref{fig:Nutshell_Overview}. Our model consists of three modules: i) joint salient edge and saliency detection module, ii) multi-scale deep feature and handcrafted feature integration module, and iii) context module.

Firstly, we take advantage of the salient edge to guide saliency detection by reformulating saliency detection as a three-category dense labeling problem (background, salient edge and salient objects) in contrast to existing saliency detection methods. Our saliency maps highlight salient objects inside salient edges. With saliency edges obtained from our model, we recover much sharper salient object edges compared with existing deep saliency methods. In addition, since our edge extraction is scale-aware, our edge-aware saliency model can better handle scenarios with small salient objects.

Secondly, inspired by CASENet \cite{CASENet}, we perform feature extraction instead of feature classification at lower stages of our deep network to suppress non-salient background region and accentuate salient edges. We integrate multi-scale deep features and handcrafted features to have more representative saliency features. Thus, we exploit the complementary nature of handcrafted and deep saliency methods in a multi-modal fashion \cite{Multimodal} and also achieve multi-scale saliency detection \cite{TinyFace}. Finally, we employ a context module \cite{Dilation-Convolution} with dilated convolution to explore global and local contexts for saliency, leading to  more accurate saliency maps with sharper edges.

Our deep edge-aware saliency model is trained by taking the responses of existing handcrafted saliency (robust background detection \cite{Background-Detection:CVPR-2014} in particular) and normalized RGB color images as inputs, and directly learning an end-to-end mapping between the inputs and the corresponding saliency maps and salient edges. Deep supervision has been enforced in the network learning.

Our main contributions can be summarized as:
\begin{enumerate}
\item We reformulate saliency detection as a three-category dense labeling problem and propose a unified framework to jointly learn saliency map and salient edges in an end-to-end manner.
\item We integrate deep features and handcrafted features into a deep-shallow model to make the best use of complementary information in data-driven deep saliency and human knowledge driven saliency.
\item We perform feature extraction instead of feature classification at earlier stages of our network to suppress non-salient pixels and provide higher-fidelity edge localization with structure information for saliency detection.
\item We use multi-scale context to exploit both global and local contextual information to produce maps with much sharper edges.
\item Our extensive performance evaluation on 10 benchmarking datasets show the superiority of our method compared with all current state-of-the-art methods.
\end{enumerate}

The remainder of the paper is organized as follows. Section \ref{sec:related_works} presents related work and summarizes the uniqueness of our method. In Section \ref{sec:joint_edge_saliency}, we introduce our three-category dense labeling formulation and a new balanced loss function to achieve edge-aware saliency detection. Deep and handcrafted features integration as well as context module for saliency refinement are explained in Section \ref{sec:deep_hand_integration}. Experimental results, performance comparisons and ablation studies are reported in Section \ref{sec:experimental_results}. Section \ref{sec:conclusion} concludes the paper with possible future work.


\section{Related Work}
\label{sec:related_works}

Saliency detection approaches can be roughly classified as handcrafted feature based methods and deep learning based methods. After an overview of these categories, we explore methods for multi-scale feature fusion in this section.

\subsection{Handcrafted Features for Saliency Detection}

Prior to the deep learning revolution, conventional saliency methods were mainly relied on handcrafted features \cite{Frequency-tuned:CVPR-2009}, \cite{Patch-Distinct:CVPR-2013}, \cite{Soft-Image-Abstraction:ICCV-2013}, \cite{Saliency-Filters:CVPR-2012}, \cite{Global-Contrast:CVPR-2011}. We refer readers to \cite{Salient-Detection-Survey:2014} and \cite{SalObjBenchmark_Tip2015} for in-depth surveys and benchmark comparisons of handcrafted saliency methods.

Given an over-segmented image, color contrast has been exploited in \cite{Global-Contrast:CVPR-2011} and \cite{YehPR}. Liu \etal \cite{Learning-Detect-Salient:CVPR-2007} formulated saliency detection as an image segmentation problem. By exploiting the sparsity prior for salient objects, Shen and Wu \cite{Low-Rank-Recovery:CVPR-2012} solved saliency detection as a low-rank matrix decomposition problem. Objectness, which highlights the object-like regions, has also been used in \cite{Objectness:PAMI-2012}, \cite{UFO-Saliency:ICCV-2013} and \cite{Fusing-Objectness-Saliency:ICCV-2011}. Xia \etal ~ \cite{6738043} measured saliency by the sparse reconstruction residual of representing the central patch with a linear combination of its surrounding patches sampled in a nonlocal manner. Zhu \etal ~\cite{Background-Detection:CVPR-2014} presented a robust background measure, namely ``boundary connectivity'', and a principle optimization framework. As a modification to commonly used center prior, Gong \etal~\cite{SaliencyPropagation15} used the center of a convex hull to obtain strong background and foreground priors. Different from the above unsupervised methods that compute pixel or superpixel saliency directly, more recent works \cite{DRFI:CVPR-2013}, \cite{High-Dim-Color-Transform:CVPR-2014} and \cite{Contextual-cue:ICCV-2011} consider saliency detection as a regression problem, in which a classifier is trained to assign saliency value to each pixel or superpixel.

\subsection{Deep Learning Based Saliency}

The above methods are effective for simple scenes, but they become fragile for complex scenarios. Recently, deep neural networks has been adopted to saliency detection \cite{InstanceSaliency}, \cite{ChengCVPR17}, \cite{TIP}, \cite{DeepMC}, \cite{DISC}, \cite{LEGS}, \cite{MDF:CVPR-2015}, \cite{RFCN}, \cite{RACDNN}, \cite{ELD}, \cite{DC}, \cite{SpCNN}. Deep networks can encode high-level semantic features that capture saliency information more effectively than handcrafted features, and report superior performance compared with the conventional techniques.

Deep learning based saliency detection methods generally train a deep neural network to assign saliency to each pixel or superpixel. Li and Yu \cite{MDF:CVPR-2015} used learned features from an existing CNN model to replace the handcrafted features. Li \etal \cite{TIP} proposed a multi-task learning framework to saliency detection, where saliency detection and semantic segmentation are learned jointly. In DISC \cite{DISC}, a novel deep image saliency computing framework is presented for fine-grained image saliency computing, where two stacked DCNNs are used to get coarse-level and fine-grained saliency map respectively. Nguyen and Liu \cite{Semantic_Saliency17} integrated semantic priors into saliency detection.Kuen \etal \cite{RACDNN} proposed a recurrent attentional convolutional-deconvolution network (RACDNN) to iteratively select image sub-regions to perform saliency refinement. Liu and Han \cite{Liu_2016_CVPR} proposed an end-to-end deep hierarchical saliency network (DHSNet). The work is similar to \cite{Pan_2016_CVPR}, where a shallow and a deep convolutional network are trained respectively in an end-to-end architecture for eye fixation prediction. In ELD \cite{ELD}, both CNN features and low-level features are integrated for saliency detection. Zhao \etal ~\cite{DeepMC} trained a local estimation stage and a global search stage individually to predict saliency score for salient object regions. Very recently, Hu \etal \cite{levelset} proposed a level-set function and a superpixel based guided filter to refine the saliency maps. Wang \etal ~\cite{imagesaliency} predicted saliency based on image level label of a given input image.

\subsection{Multi-scale Feature Fusion}

Fusion of features in multiple spatial scales has been shown as a factor in achieving the state-of-the-art performance on semantic segmentation \cite{attentionmodel} \cite{PSPNet} \cite{Dilation-Convolution} and edge detection \cite{HED15}. As an extension of \cite{MDF:CVPR-2015}, Li and Yu \cite{MDF_TIP} added handcrafted features to the deep features and trained a random forest (MDF-TIP) to predict saliency for each superpixel. Wang \etal~\cite{RFCN} proposed a recurrent neural network, which takes unsupervised saliency and RGB image as input, and recurrently update output saliency map. Li and Yu \cite{DC} used an end-to-end contrast network (DC) to produce a pixel-level saliency map based on multi-level features from four bottoms pooling layers based on the VGG network \cite{VGG}. Within the structure of HED \cite{HED15}, Cheng \etal ~\cite{ChengCVPR17} proposed a method (DSS) by introducing short connections to the skip-layer structures of \cite{HED15}. UberNet \cite{UberNet} uses a unified architecture to jointly handle low-, mid- and high-level vision task based on a skip architecture at deeply supervision manner, where features from bottom layers and top layers are fused for different dense prediction tasks.


In contrast to the above state-of-the-art deep saliency methods, especially those based on multi-level feature fusion ~\eg~ MDF-TIP \cite{MDF_TIP}, RFCN \cite{RFCN}  DSS \cite{ChengCVPR17}, DC \cite{DC} and UberNet \cite{UberNet}, our fully convolutional framework can efficiently aggregate saliency cues across different layers by using feature extraction instead of feature classification. Furthermore, by incorporating handcrafted saliency maps as a part of our input, our model can utilize the statistical cues and also initiate its parameters with reasonable weights. At the same time, by reformulating saliency detection as a three-category dense-labeling problem, we build our model in an end-to-end manner and generate a spatially dense and highly coherent, edge-guided saliency map.

\section{Joint Salient Edge and Saliency Detection}
\label{sec:joint_edge_saliency}


To highlight salient edges as well as better detect salient objects of diverse scales, we reformulate saliency detection as a three-category dense labeling task and jointly learn salient edge and saliency map within our end-to-end framework. Furthermore, to handle the imbalance in labeling, we propose a new loss function.


\subsection{Reformulating Saliency Detection}

In our three-category labeling scheme, we introduce a new label ``salient edge''. Under our formulation, saliency detection aims at labeling each pixel with one of the three categories, namely background, salient edge and salient objects.
\begin{figure}[!htp]
   \begin{center}
   \begin{tabular} {c@{ }  c@{ } c@{ } }
   {\includegraphics[width=0.32\linewidth]{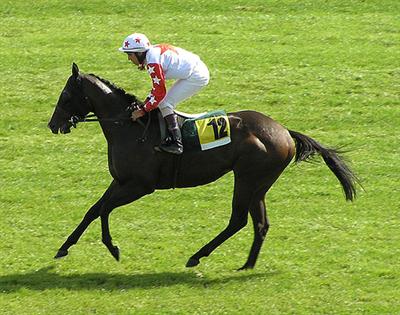}} &
   {\includegraphics[width=0.32\linewidth]{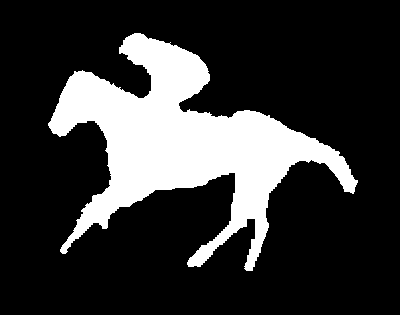}}&
   {\includegraphics[width=0.32\linewidth]{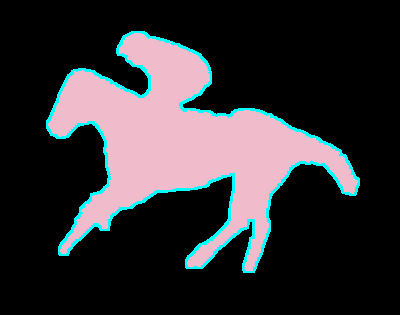}} \\
   \footnotesize{(a) Image} & \footnotesize{(b) Standard label} & \footnotesize{(c) Our new label}
   \end{tabular}
   \end{center}
   \vspace{-2mm}
   \caption{Illustration of relabeled ground truth. Our new saliency ground truth consists of three categories, namely background (black), saliency edge (blue) and salient object region (pink).}
   \label{fig:Relabel_ground_truth}
\end{figure}

Since all existing saliency benchmark datasets have two-category labels, we need to transform the available two-category labels to our three-category version. Conventionally, for each image $I$, its ground truth saliency is denoted as $G = \{l_b, l_s\}$, where the background region $l_b$ is labeled in black, and salient object region $l_s$ is labeled in white. To recover the location information of salient object, we convert labels of the saliency detection dataset into three categories: 1) background $l_b$, 2) salient edge $l_{e}$, and 3) salient objects $l_s$.

Given a ground-truth saliency map $G$, we use the ``Canny'' edge detector to extract an initial edge map $E$. As the edge map $E$ tends to be very thin (1 or 2-pixel width), a dilation operation is performed. For each pixel in $E$, we label its $3 \times 3$ neighboring region as edge region. Thus we end up with thicker edge regions. We generate new labels of the edge guided saliency in the following way: 1) Label all pixels as background; 2) Assign salient edge label to the thickened edge region; 3) Assign salient object label to salient object region.

In this way, completeness of salient object is kept as well as the accuracy of salient object edge. We compare the two-category and three-category labeling strategies in Fig.~\ref{fig:Relabel_ground_truth}, where the salient edge has been greatly emphasized.

\subsection{Balanced Loss Function}
\label{subsec:loss_function}


For typical images, the distribution of the background/salient edge/salient region pixels is heavily imbalanced; generally around 90\% of the whole image is labeled as background or salient objects while less than 10\% of the entire image is labeled as salient edges. Inspired by HED \cite{HED15} and InstanceCut \cite{Instancecut}, we propose a simple yet effective strategy to automatically compensate the loss among the three categories by introducing a class-balancing weight on a per-pixel basis.

Specifically, we define the following class-balanced softmax loss function in Eq.~\eqref{eq:BalancedLoss}:
\begin{equation}
\begin{aligned}
Loss = -\beta_b\sum_{j\in Y_b}\log P_r(y_j = 0)-\beta_e\sum_{j\in Y_e}\log P_r(y_j = 1)-\\
\beta_s\sum_{j\in Y_s}\log P_r(y_j = 2),
\label{eq:BalancedLoss}
\end{aligned}
\end{equation}
where $j$ indexes the whole image spatial dimension, $\beta_b = (\vert Y_e \vert+\vert Y_s\vert) /\vert Y \vert $, $\beta_e = (\vert Y_b \vert+\vert Y_s\vert) /\vert Y \vert $, $\beta_s = (\vert Y_b \vert+\vert Y_e\vert) /\vert Y \vert $. $\vert Y_b \vert$, $\vert Y_e \vert$, $\vert Y_s \vert$ and $\vert Y \vert$ denote pixel number of the background,
\begin{figure}[!htp] \small
   \begin{center}
   \begin{tabular} {c@{ }  c@{ } c@{ } c@{ } c@{ }}
   {\includegraphics[width=0.18\linewidth]{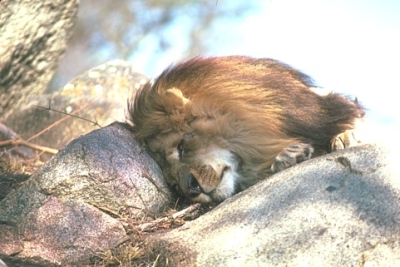}} &
   {\includegraphics[width=0.18\linewidth]{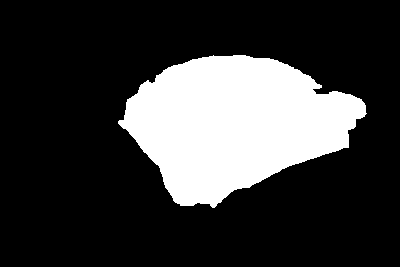}} &
   {\includegraphics[width=0.18\linewidth]{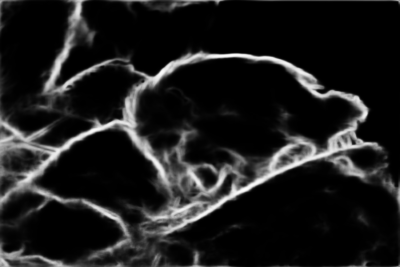}} &
   {\includegraphics[width=0.18\linewidth]{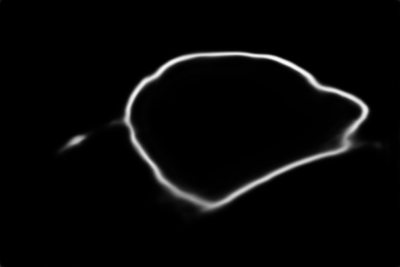}} &
   {\includegraphics[width=0.18\linewidth]{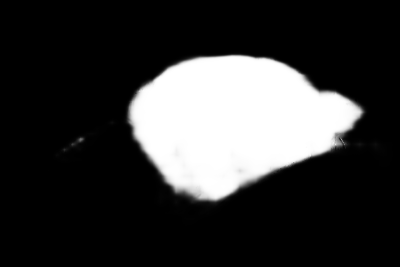}} \\
   \footnotesize{Image} & \footnotesize{GT} & \footnotesize{Edge by \cite{HED15}} & \footnotesize{Salient Edge} & \footnotesize{Salient Object}
   \end{tabular}

   \end{center}
   \vspace{-2mm}
   \caption{Edge maps produced by our model and HED \cite{HED15}.}
   \label{fig:hed_ouredge}
\end{figure}
salient edge, salient object and the whole image respectively. $P_r(y_j = 0) = e^{b_j}/\sum_n e^{b_n}\in [0,1]$ is computed using the Softmax function at pixel $j$ which represent the possibility of pixel $j$ to be labeled as background. Similarly, $P_r(y_j = 1) = e^{b_j}/\sum_n e^{b_n}\in [0,1]$ and $P_r(y_j = 2) = e^{b_j}/\sum_n e^{b_n}\in [0,1]$ denote the possibility of pixel $j$ labeled as salient edge and salient object respectively.


\subsection{FCN for Edge-Aware Saliency}
\label{subsec:fcns_edge}

Our front-end saliency detection network is built upon a semantic segmentation net, \ie, DeepLab \cite{Deeplab}, where a deep convolutional neural network (ResNet-101 \cite{ResHe2015}) originally designed for image classification is repurposed to the task of semantic segmentation by 1) transforming all fully connected layers to convolutional layers, 2) increasing feature resolution through dilated convolutional layers \cite{Deeplab}, \cite{Dilation-Convolution}. Under this framework, the spatial resolution of the output feature map is increased four times, which is superior to \cite{DeepMC} and \cite{MDF:CVPR-2015}.

Different from VGG \cite{VGG}, ResNet \cite{ResHe2015} explicitly learns residual functions with reference to the layer inputs, which makes it easier to optimize with higher accuracy from considerably increased network depth. By removing the final pooling and fully-connected layer to adapt it to saliency detection, we reconstruct ResNet-101 model, and add four dilated convolutional layers with increasing receptive field in our saliency detection network to better exploit both local and global context information.

For a normalized RGB image $I$, with the repurposed ResNet-101 model, we get a three-channel feature map $S_{ deep} = \{S_{b},S_{e},S_{s}\}$, where the first channel represents background map, second channel salient edge map and third channel salient object map. We up-sample $S_{ deep}$ to the input image resolution, and compute loss and accuracy using the interpolated feature map. Note that, this up-sampling operation could also be achieved by deconvolution \cite{DeconvCNN}), which leads to more parameters and longer training time with similar performance. In this paper, we use the ``Interp'' layer provided in Caffe \cite{jia2014caffe} due to its efficiency.

\subsection{Effects of Reformulating Saliency Detection}

We compare salient edge from our model and edge map from state-of-the-art deep edge detection method HED \cite{HED15}. As shown in Fig.~\ref{fig:hed_ouredge}, our salient edge map provides rich semantic information, which highlights the salient object edges and suppresses most of the background edges.

\begin{figure*}[!htp]
   \begin{center}
   \begin{tabular} {c c c}
{\includegraphics[width=0.32\linewidth]{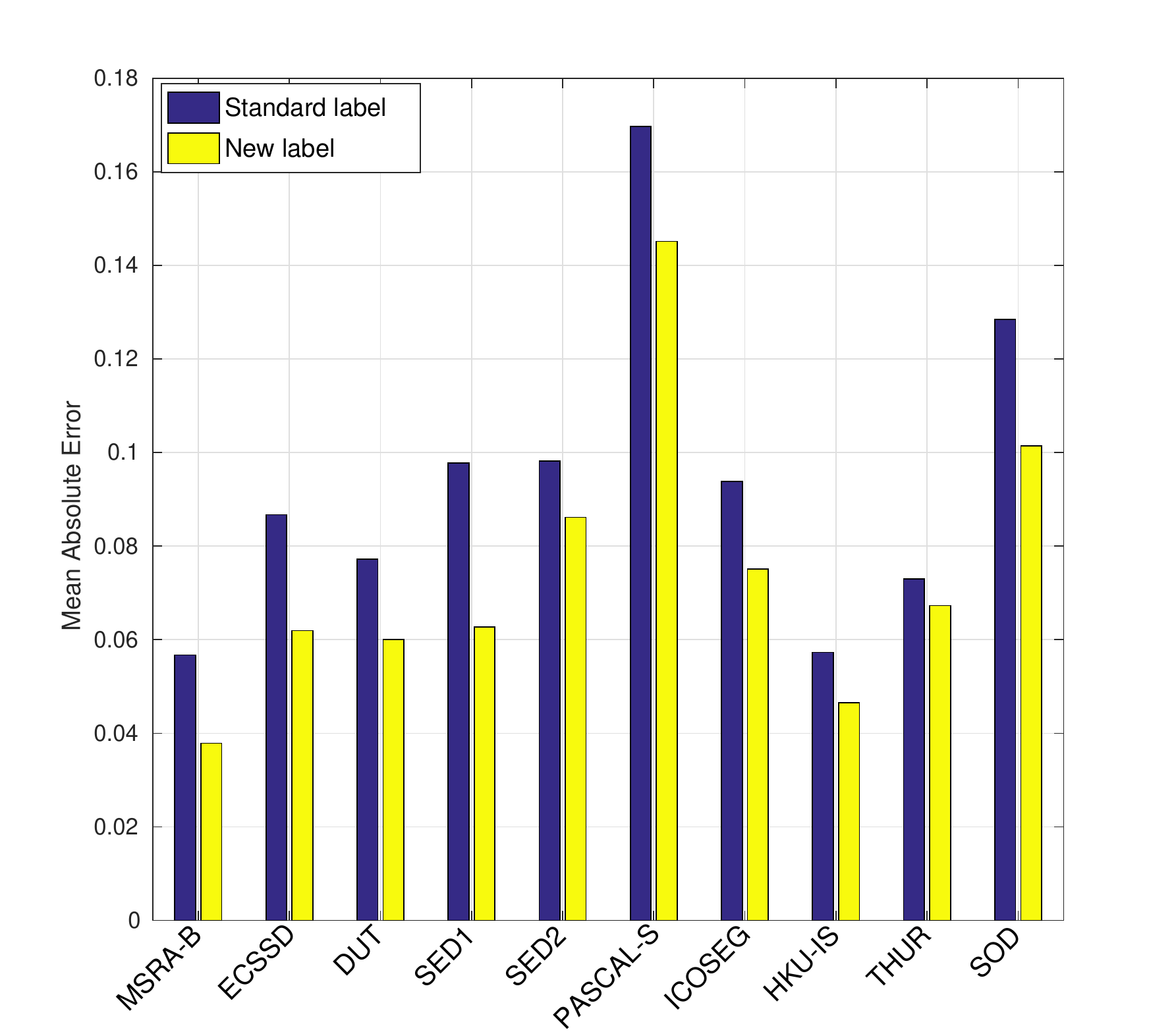}} &
   {\includegraphics[width=0.32\linewidth]{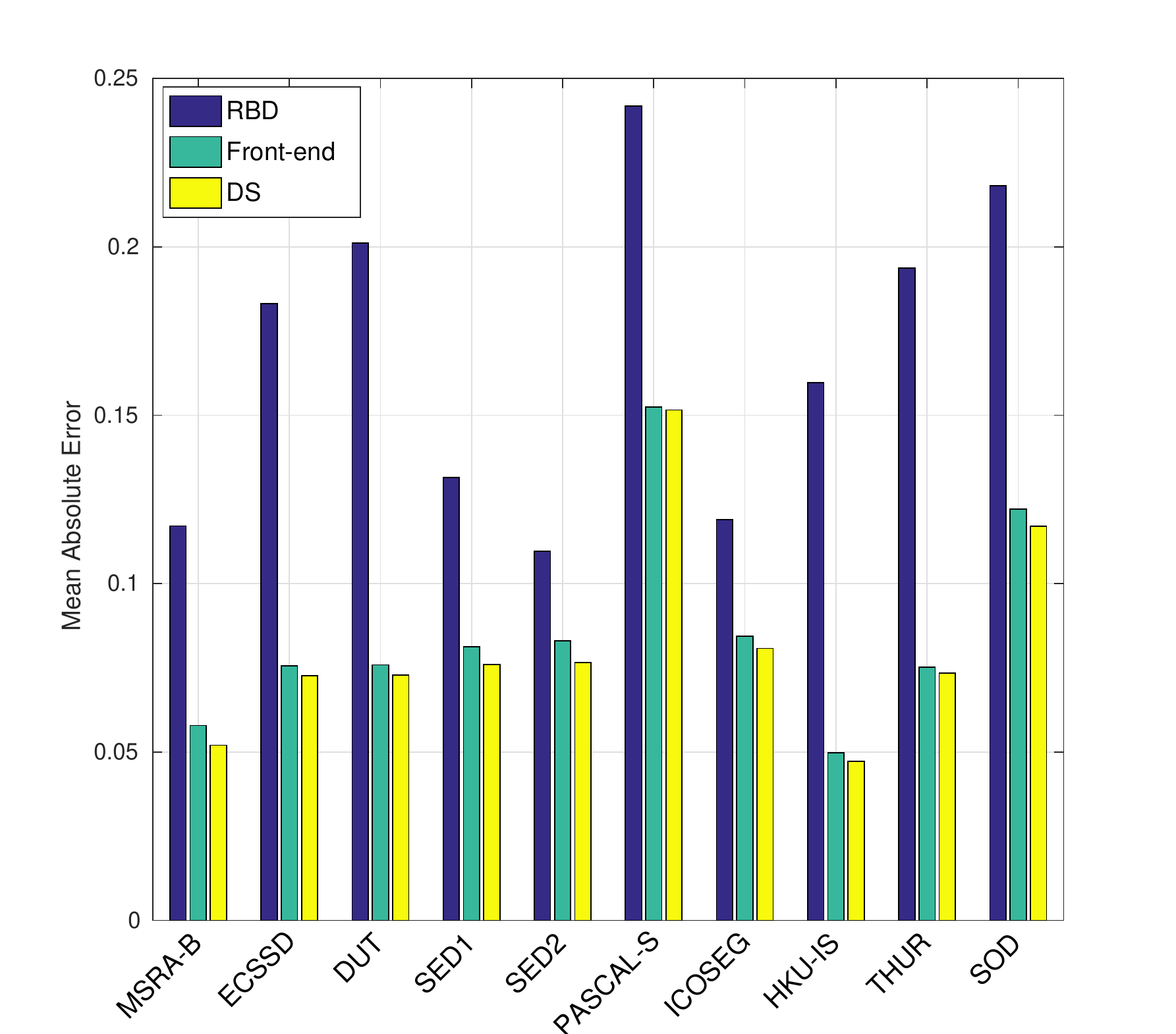}} &
   {\includegraphics[width=0.32\linewidth]{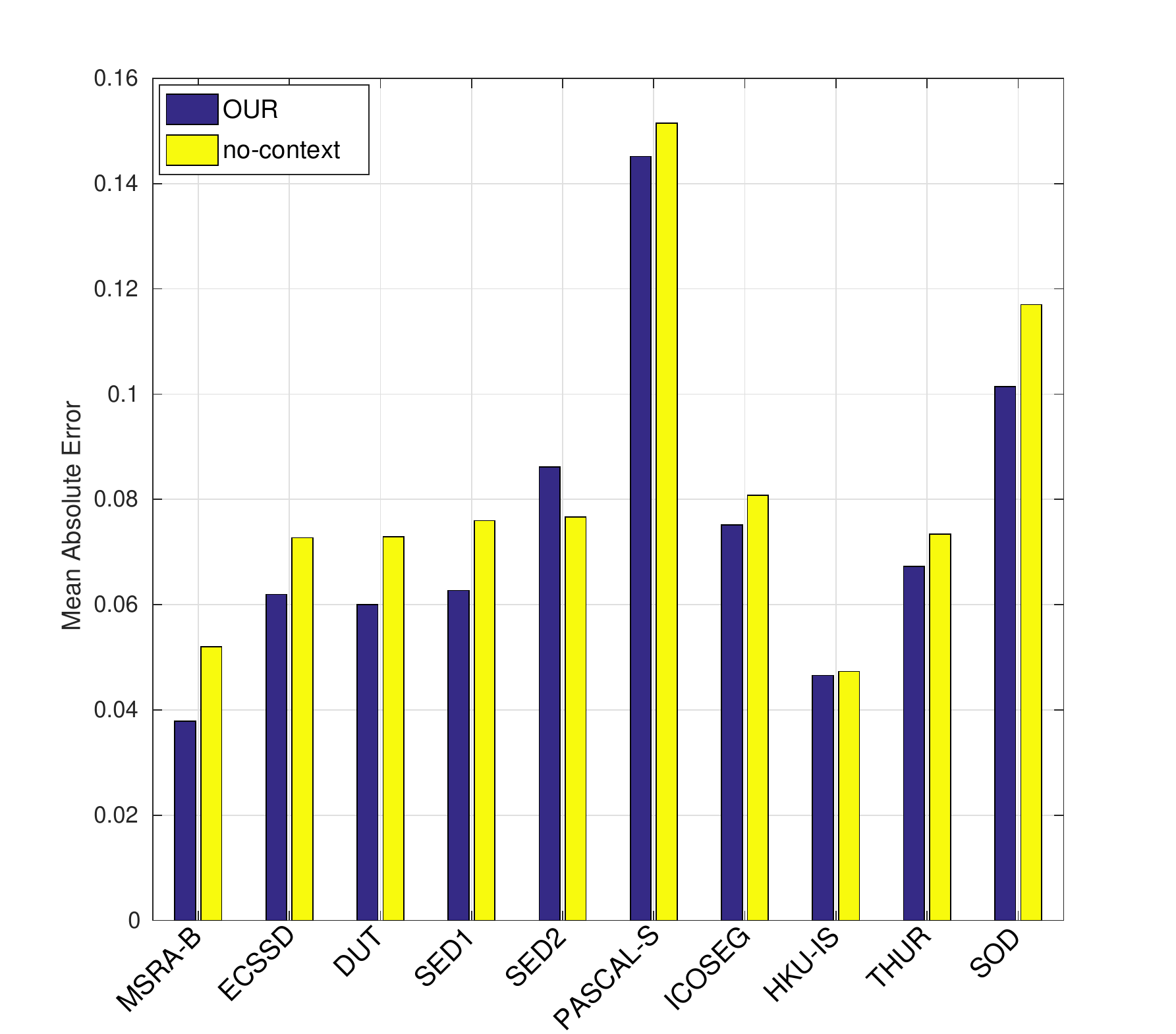}} \\
  \footnotesize{(a) Reformulating saliency detection}  & \footnotesize{(b) Deep-handcrafted feature integration} & \footnotesize{(c) Context module refinement} \\
    \end{tabular}
   \end{center}
   \vspace{-2mm}
   \caption{Mean Absolute Error (MAE) on 10 benchmark datasets for models analysis. (a) Performance difference of using the standard two-category labeling and using our reformulated three-category labeling. Models in (b) are all based on three-category labeling, which illustrates how deep and unsupervised feature integration helps the performance of our model. (c) Two models of using and not using context module.}
   \label{fig:method_analysis}
\end{figure*}

To analyze the importance of our new saliency detection formulation, we train an extra deep model, with two-category labeling as ground truth, namely ``Standard label'', and our model using the relabeled ground truth is named as ``New label''. For the above two models, we use similar model structure, and the only difference is the channel number of outputs (with the standard label as ground truth, ``num\_output: 2'' and our new label as ground truth ``num\_output: 3''). Fig.~\ref{fig:method_analysis} (a) shows the mean absolute error (MAE) on ten benchmarking datasets, which clearly demonstrates  that with our new labeling strategy, we end up with consistently better performance on all the ten benchmarking datasets, with approximate 2.5\% decrease in MAE on average, which clearly proves the effectiveness of our reformulation of saliency detection.


\section{Integrating Deep and Handcrafted Features}

\label{sec:deep_hand_integration}
Deep learning based saliency detection methods predict saliency maps by exploiting large scale labeled saliency datasets, which could be biased by the training datasets. By contrast, handcrafted saliency detection methods build upon statistical priors that are summarized and extracted with human knowledge, and are more generic and applicable to general cases. Current deep learning networks generally neglect the statistical priors widely used in handcrafted saliency methods.

Here, we propose to integrate both deep saliency from our front-end deep model and handcrafted saliency for edge-aware saliency detection. We choose RBD (Robust Background Detection) \cite{Background-Detection:CVPR-2014} as our handcrafted saliency model, as it ranks 1st of all the handcrafted saliency detection methods \cite{SalObjBenchmark_Tip2015}. We add a shallow model (as shown in Fig.~\ref{fig:Nutshell_Overview}) to fuse deep and handcrafted features, which takes original RGB image $I$, handcrafted saliency $S_{RBD}$, saliency map $S_{deep}$ from our front-end model and lower levels feature map $S_{i}, i = 1,...,4$ from our deep model as input. We train the feature integration model to extract complementary information between deep and handcrafted features. Finally, we plug a context module by using dilated convolution at the end of our feature fusion model to generate saliency map with sharper edges.

\subsection{Handcrafted Saliency Features}
\label{subsec:unsupervised_saliency}

We select RBD \cite{Background-Detection:CVPR-2014} as our handcrafted saliency detection model. RBD is developed based on the statistical observation that objects and background regions in natural images are quite different in their spatial layout, and object regions are much less connected to image boundaries than background ones. Background connectivity is defined to measure how much a given region is connected to image boundaries:
\begin{equation}
BndCon(p)=\frac{Len_{bnd}(p)}{\sqrt[]{Area(p)}},
\end{equation}
where $Len_{bnd}(p)$ and $Area(p)$ are length along the image boundary and spanning area of superpixel $p$ respectively. Based on this measure, background probability $\omega_i^{bg}$ is defined as below, which is close to 1 when boundary connectivity is large and close to 0 when it is small:
\begin{equation}
\omega_i^{bg} =1-\exp\left(-\frac{BndCon^2(p_i)}{2\delta^2_{bndCon}}\right).
\label{eq:background_probability}
\end{equation}
Furthermore, background weighted contrast is introduced to compute saliency for a given region $p$, which is defined as:
\begin{equation}
wCtr(p) =\sum_{i=1}^N d_{app}(p,p_i)\omega_{spa}(p,p_i)\omega_i^{bg},
\label{eq:weighted_contrast}
\end{equation}
where $d_{app}(p,p_i)$ is the Euclidean distance between average colors of region $p$ and $p_i$ in the CIE-Lab color space, $\omega_{spa}(p,p_i)$ is the spatial distance between the center of region $p$ and $p_i$. Please refer to \cite{Background-Detection:CVPR-2014} for more details.

RBD has a clear geometrical interpretation, which makes the boundary connectivity robust to image appearance variations and stable across different images. Furthermore, the saliency cues captured in RBD have strong correlation with the geometric coordinates and weak correlation with appearance, which is essentially different from data-driven deep saliency methods. In Fig.~\ref{fig:saliency_compare1}, we illustrate examples where RBD could capture the salient objects while state-of-the-art deep learning based methods fail to localize the salient objects.

\subsection{Deep Multi-scale Saliency Feature Extraction}
\label{subsec:deep_feature_extraction}

\begin{figure}[!htp]
   \begin{center}
   \begin{tabular} {c@{ }  c@{ } c@{ } c@{ } c@{ }}
   {\includegraphics[width=0.18\linewidth]{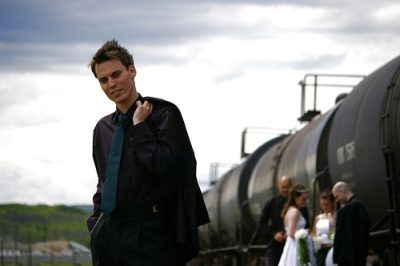}} &
   {\includegraphics[width=0.18\linewidth]{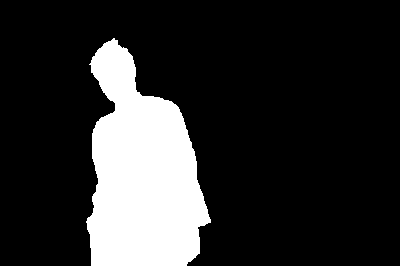}} &
   {\includegraphics[width=0.18\linewidth]{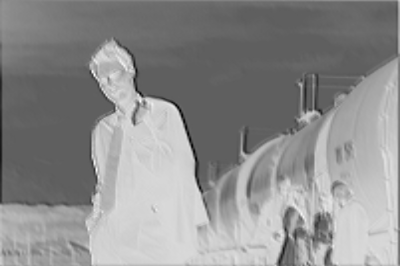}} &
   {\includegraphics[width=0.18\linewidth]{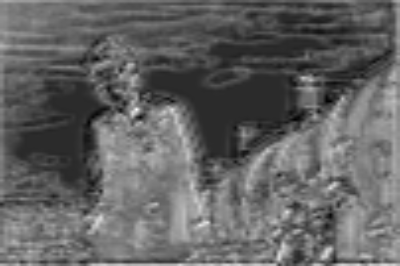}} &
   {\includegraphics[width=0.18\linewidth]{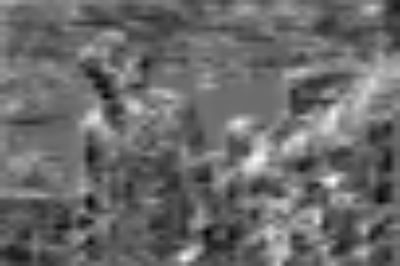}} \\
    \small{Image} &\small{GT} &\small{$conv1$} &\small{$res2c$} &\small{$res3b3$} \\
   {\includegraphics[width=0.18\linewidth]{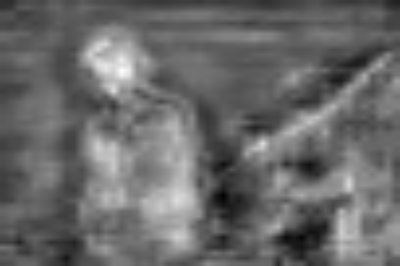}} &
   {\includegraphics[width=0.18\linewidth]{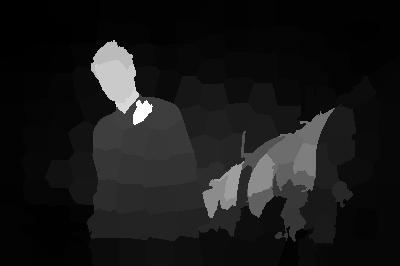}} &
   {\includegraphics[width=0.18\linewidth]{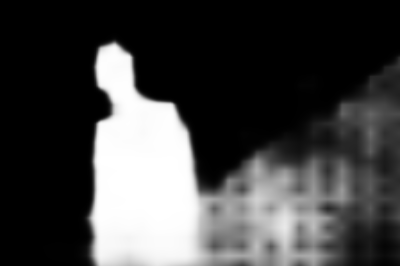}} &
   {\includegraphics[width=0.18\linewidth]{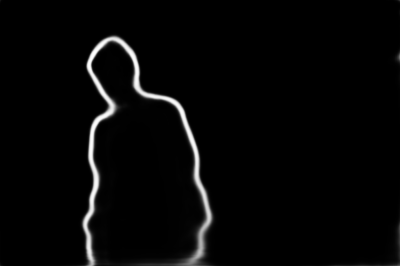}} &
   {\includegraphics[width=0.18\linewidth]{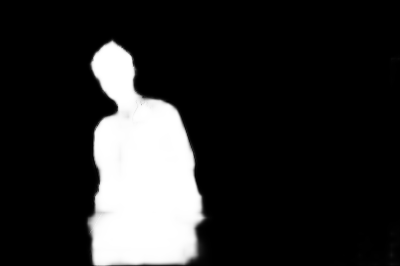}} \\
   \footnotesize{$res4b22$} &\footnotesize{RBD} &\footnotesize{$S_{s}$} &\footnotesize{Edge} &\footnotesize{Ours} \\
      \end{tabular}
   \end{center}
   \vspace{-2mm}
   \caption{Illustration of how deep and handcrafted feature integration helps the performance of our model. Salient region in this image is in low contrast, and output of our front-end deep model ($S_s$) failed to suppress some background region. With help of the deep-handcrafted integrated feature, we end up with clear salient edge (``Edge'') and saliency map (``Ours'').}
   \label{how_shallow_helps}
\end{figure}

It has been shown that lower layers in convolutional nets capture rich spatial information, while upper layers encode object-level knowledge but are invariant to factors such as pose and appearance \cite{SharpMask}. In general, the receptive field of lower layers is too limited, it is unreasonable to require the network to perform dense prediction at early stages. To this end, inspired by CASENet \cite{CASENet}, we perform feature extraction at the last layer of each block of the re-purposed ResNet-101 ($conv1$, $res2c$, $res3b3$ and $res4b22$ respectively in our paper). Particularly, one $1\times 1$ convolution layer is utilized to map each of the above four side-outputs to one-channel feature map $S_{i}, i = 1,...,4$, which represents different scales of detail information of our deep model, see Fig.~\ref{how_shallow_helps} for an example, where $S_{s}$ from our front-end model is trained without deep-handcrafted feature fusion.

Figure~\ref{how_shallow_helps} shows that bottom side-outputs are usually messy which makes them undesirable to infer dense prediction at these stages. However, as lower level outputs can produce more detail information, especially edges, it will be a huge waste to ignore them at all. In this paper, we take a trade-off between keeping detail information and suppressing non-salient object pixels by performing feature extraction instead of feature classification at earlier stages. Those extracted low-level features can provide spatial information for our final results at the top layer, which will be discussed later.


\subsection{Deep-Shallow Model}
\label{subsec:integrating_deep_unsupervised}

With saliency map $S_{RBD}$ from \cite{Background-Detection:CVPR-2014}, input image $I$ and output from our front-end deep model $S_{deep} = \{S_{b},S_{e},S_{s}\}$, as well as the four side-outputs $S_{i}, i = 1,...,4$, we train a three-layer shallow convolutional networks to better explore statistical information as well as to extract complementary information between deep saliency and handcrafted saliency.

Firstly, we concatenate $I$, $S_{s}$, $ S_{RBD}$ and $S_{i}, i = 1,...,4$ in channel dimension and feed it to our three-layer convolutional model to map the concatenated feature map to a one-channel feature map $S_{ns}$. Then, background map $S_{b}$, edge map $S_{e}$ and $S_{ns}$ are concatenated and fed to another $1\times 1$ convolution layer to form our edge-aware saliency map $S_{DS}$ with a loss layer at the end. Although deep learning based method outperform RBD with a margin, and feature map from the earlier stages are undesirable, experimental results on ten benchmark datasets demonstrate that the proposed model to integrate handcrafted saliency and deep saliency from bottom layers achieves better performance, see Fig.\ref{how_shallow_helps} for an example.


To illustrate how deep-handcrafted saliency integration helps the performance of our model, we compute MAE on ten datasets as shown in Fig.~\ref{fig:method_analysis}(b), where ``RBD'' represents the performance of using handcrafted RBD \cite{Background-Detection:CVPR-2014}, ``Front-end'' represents the performance by training our deep model without deep-handcrafted saliency, and ``DS'' represents results from our proposed deep-shallow model. As shown in Fig.~\ref{fig:method_analysis}(b), with the help of deep-handcrafted saliency integration, our model achieves consistently lower MAE. With the detail information of deep features from bottom sides and statistical information from RBD, our model is effective in detecting salient objects of diverse scales, and saliency map of our model becomes sharper and better visually as shown in Fig.~\ref{fig:sample_show}.

\subsection{Context Module for Saliency Refinement}
\label{subsec:context_module}

\begin{figure}[!htp]
   \begin{center}
   \begin{tabular} {c@{ }  c@{ } c@{ }}
   {\includegraphics[width=0.316\linewidth]{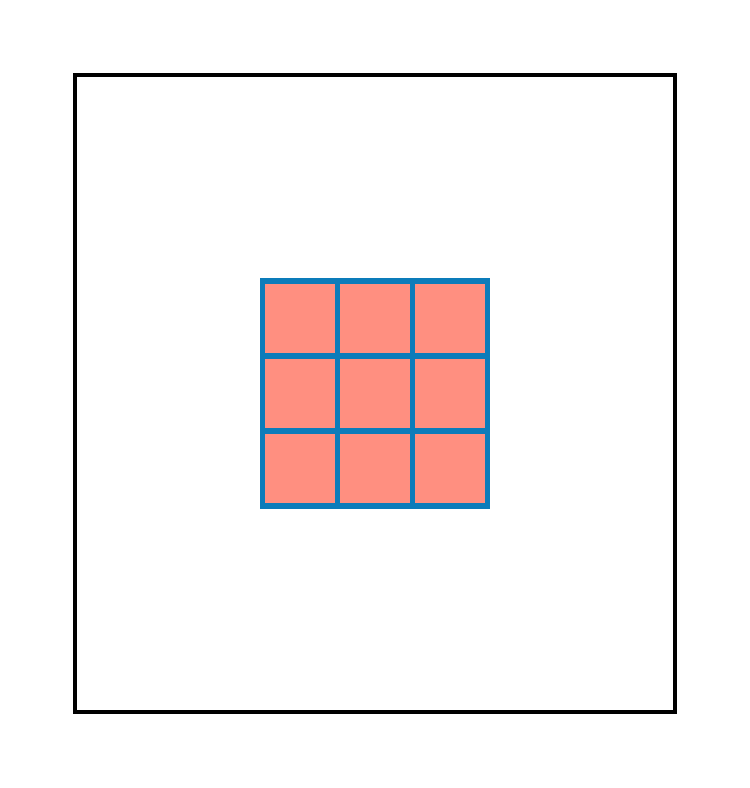}} &
   {\includegraphics[width=0.316\linewidth]{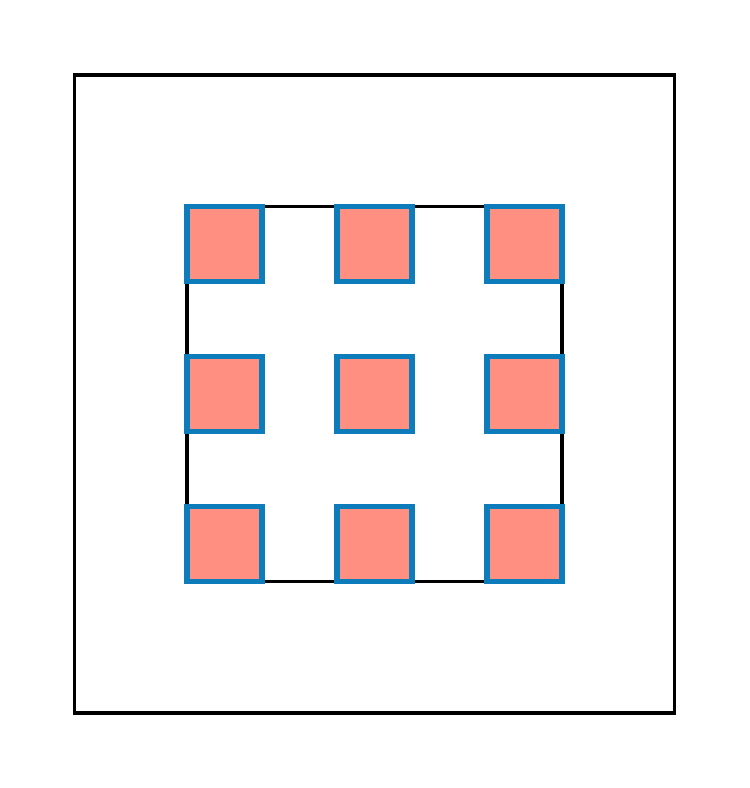}} &
   {\includegraphics[width=0.30\linewidth]{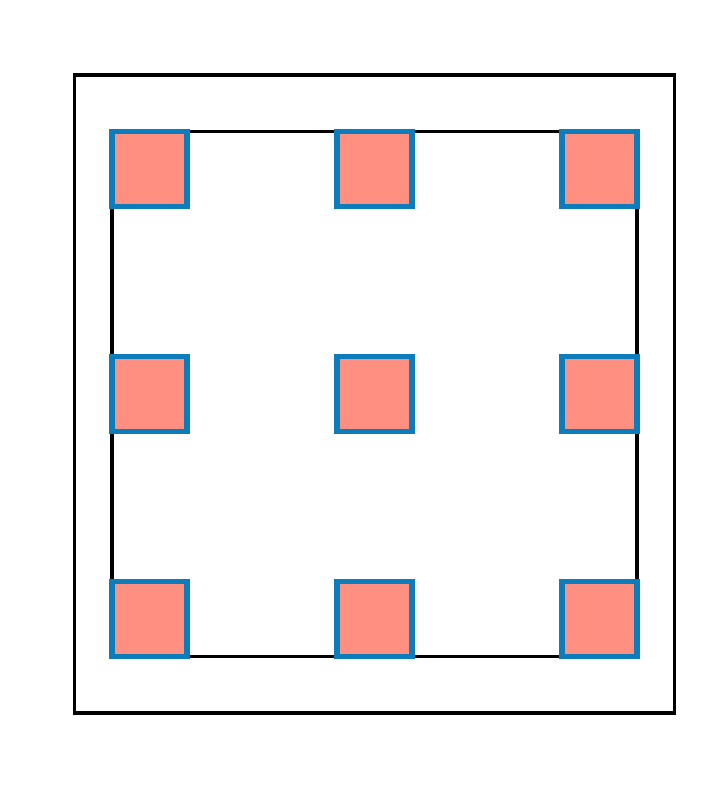}} \\
   \footnotesize{(a) pad=1,dilation=1} & \footnotesize{(b) pad=2,dilation=2} & \footnotesize{pad=4,dilation=4} \\
   \end{tabular}
   \end{center}
   \vspace{-2mm}
   \caption{Illustration of 2D dilated convolution. (a) is produced from a 1-dilated convolution, with receptive field of $3\times3$. (b) is produced from (a) by a 2-dilated convolution with receptive field of $7\times7$. (c) is produced from (b) by a 4-dilated convolution with receptive field of $15\times15$.}
   \label{fig:dilation}
\end{figure}

Using our deep-shallow model, we produce a dense saliency map. To further improve edge sharpness of the saliency map, a context module \cite{Dilation-Convolution} is added at the end of our deep-shallow network. By systematically applying dilated convolutions \cite{Dilation-Convolution} for multi-scale context aggregation, context network could exponentially expand the receptive fields without losing resolution or coverage as shown in Fig.~\ref{fig:dilation}, where a fixed size of $kernal=3$ is applied with different scale of dilation.

Let $F_i,i=0,1,...,n-1:\mathbb{Z}^2 \to \mathbb{R}$ be discrete functions and let $k_j,j=0,1,...,n-2:\Omega_r \to \mathbb{R}$ be discrete $(2j+1)\times(2j+1)$ filters. Dilated convolution \cite{Dilation-Convolution} is defined as:
\begin{equation}
(F*_lk)(p) = \sum_{s+lt=p}{F(s)k(t)},
\label{eq:dialted_conv_defination}
\end{equation}
where $*_l$ is an $l$-dilated convolution, $p$, $s$ and $t$ are elements in $F*_l$, $F$ and $k$ respectively. Given $F_i$ and $k_i$, consider applying the filters with exponentially increasing dilation:
\begin{equation}
F_{i+1}=F_i*_{2^i}k_i \quad for\ i=0,1,...,n-2.
\label{eq:dialted_conv}
\end{equation}
It's easy to observe that the receptive filed of elements in $F_{i+1}$ is $(2^{i+2}-1)\times(2^{i+2}-1)$, which is a square of exponentially increasing size.　See \cite{Dilation-Convolution} for details about context module.

\begin{figure}[htp]
   \begin{center}
   \begin{tabular}{c@{ }  c@{ } c@{ } c@{ }}
   {\includegraphics[width=0.23\linewidth]{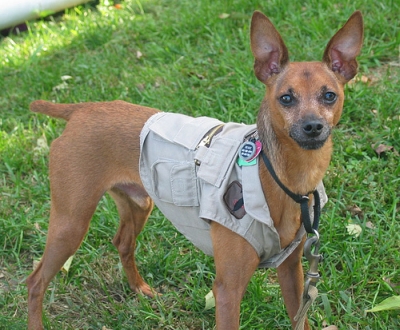}}&
   {\includegraphics[width=0.23\linewidth]{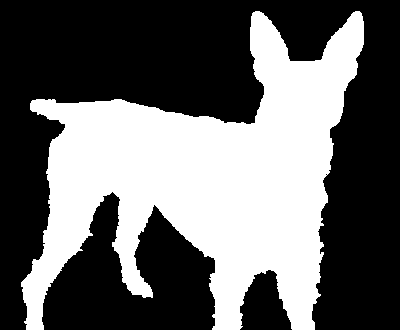}}&
   {\includegraphics[width=0.23\linewidth]{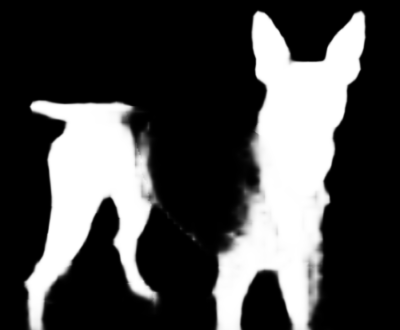}}&
   {\includegraphics[width=0.23\linewidth]{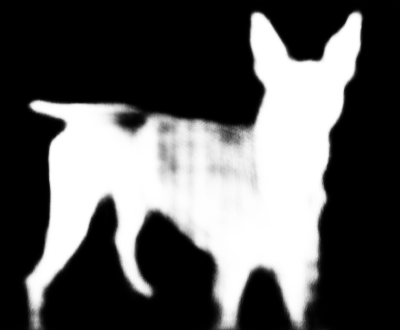}} \\
   \footnotesize{Image} & \footnotesize{GT} &\footnotesize{no-context} &\footnotesize{OUR}
      \end{tabular}
   \end{center}
   \vspace{-2mm}
   \caption{Illustration of how context module help the performance of our model. From left to right: Input image, ground truth saliency map, saliency map without and with context module.}
   \label{fig:context_module_effect}
\end{figure}

Here, we used the larger context network that uses a larger number of feature maps in the deeper layers as it achieves better performance for image semantic segmentation \cite{Dilation-Convolution}. Identity initialization \cite{Dilation-Convolution} is used to initialize the context module, where all the filters are set that each layer simply passes the input directly to the next.
With this initialization scheme, the context model helps produce saliency map with even clear semantics. Fig.~\ref{fig:context_module_effect} shows example saliency maps with and without using context module. We can conclude that with the help of context module, the increasing size of receptive field helps to produce more coherent saliency map.

To illustrate how context module improve our performance, we train models with and without context module, and results are reported in Fig.~\ref{fig:method_analysis}(c), where ``no-context'' represents model without using context module. As shown in Fig.~\ref{fig:method_analysis}(c), we achieve better performance with the help of context module, with reducing of MAE on nine out of ten benchmark datasets, except for the SED2 dataset, where more than half images in this dataset are in low contract with salient objects distribute in almost the entire image area, and this attribute hinders the performance of context model.
\section{Experimental Results}
\label{sec:experimental_results}

\subsection{Setup}
\label{subsec:experimental_setup}
\textbf{Dataset:} We have evaluated the performance of our proposed model on 10 saliency benchmarking datasets.

We used 2,500 images from the MSRA-B dataset\cite{Learning-Detect-Salient:CVPR-2007} for training and 500 images for validation, and the remaining 2,000 images for testing, which is same to \cite{DRFI:CVPR-2013} and \cite{ChengCVPR17}. Most of the images in MSRA-B dataset contain one salient object. The ECSSD dataset \cite{Hierarchical:CVPR-2013} contains 1,000 images of semantically meaningful but structurally complex images, which makes it very challenging. The DUT dataset \cite{Manifold-Ranking:CVPR-2013} contains 5,168 images. The SOD saliency dataset \cite{DRFI:CVPR-2013} contains 300 images, where many images have multiple salient objects with low contrast. The SED1 and SED2 \cite{SED} datasets contain 100 images respectively, where images in SED1 contain one single salient object while images in SED2 contain two salient objects. The PASCAL-S \cite{PASCALS} dataset is generated from the PASCAL VOC \cite{PASCAL_VOC} dataset and contains 850 images. HKU-IS \cite{MDF:CVPR-2015} is a recently released saliency dataset with 4,447 images. The THUR dataset \cite{THUR} contains 6,232 images of five classes, namely ``butterfly'',``coffee mug'',``dog jump'',``giraffe'' and ``plane''. ICOSEG \cite{ICOSEG} is an interactive co-segmentation dataset, which contains 643 images with single or multiple salient objects.

\textbf{Competing methods:} We compared our method against eleven state-of-the-art deep saliency detection methods: DSS \cite{ChengCVPR17}, DMT \cite{TIP}, RFCN \cite{RFCN}, DISC \cite{DISC}, DeepMC \cite{DeepMC}, LEGS \cite{LEGS}, MDF \cite{MDF:CVPR-2015}, RACDN \cite{RACDNN}, ELD \cite{ELD}, SPCNN \cite{SpCNN} and DC \cite{DC}, and five conventional saliency detection methods: DRFI \cite{DRFI:CVPR-2013}, RBD \cite{Background-Detection:CVPR-2014}, DSR \cite{DSR_ICCV13}, MC \cite{MC_ICCV13}, and HS \cite{Hierarchical:CVPR-2013}, which were proven in \cite{SalObjBenchmark_Tip2015} as the state-of-the-art before the deep learning revolution. We have three alternate ways to obtain results of these methods. First, we use the saliency maps provided in the paper. Second, we run the released codes provided by the authors. For those methods without code and saliency maps, we use the performance reported in other papers.

\textbf{Evaluation metrics:} We use 4 evaluation metrics, including the mean absolute error (MAE), maximum F-measure, mean F-measure, as well as PR curve. MAE can provide a better estimate of the dissimilarity between the estimated and ground truth saliency map. It is the average per-pixel difference between ground truth and estimated saliency map, normalized to [0, 1], which is defined as:
\begin{equation}
MAE = \frac{1}{W\times H} \sum_{x=1}^W \sum_{y=1}^H |S(x,y) - GT(x,y)|,
\label{eq:MAR}
\end{equation}
where $W$ and $H$ are the width and height of the respective saliency map $S$, $GT$ is the ground truth saliency map.

The F-measure ($F_{\beta}$) is defined as the weighted harmonic mean of precision and recall:
\begin{equation}
F_{\beta} = (1+\beta^2)\frac{Precision\times Recall}{\beta^2 Precision + Recall},
\label{eq:F_Measure}
\end{equation}
where $\beta^2=0.3$, $Precision$ corresponds to the percentage of salient pixels being correctly detected, $Recall$ corresponds to the fraction of detected salient pixels in relation to the ground truth number of salient pixels. The Precision-Recall (PR) curves are obtained by thresholding the saliency map in the range of [0, 255]. Mean F-measure and maximum F-measure are defined as the mean and maximum of $F_{\beta}$ correspondingly, where the former one represents a summary statistic for the PR curve and the latter one provides an optimal threshold as well as the best performance a detector can achieve.

\textbf{Training details:} We trained our model using Caffe \cite{jia2014caffe}, where the training stopped when training accuracy kept unchanged for 200 iterations with maximum iteration 15,000. Each image is rescaled to $321\times 321 \times 3$. We initialized our model using the Deep Residual Model trained for semantic segmentation \cite{Deeplab}. Weights of our deep-handcrafted saliency integration model are initialized using the ``xavier'' policy, bias is initialized as constant. We used the stochastic gradient descent method with momentum 0.9 and decreased learning rate 90\% when the training loss did not decrease. Base learning rate is initialized as 1e-3 with the ``poly'' decay policy. For loss, the balanced ``Softmaxwithloss'' is utilized. For validation, we set ``test\_iter'' as 500 (test batch size 1) to cover the full 500 validation images. The whole training took 25 hours with training batch size 1 and ``iter\_size'' 20 on a PC with an NVIDIA Quadro M4000 GPU.

\subsection{Comparison with the State-of-the-art}
\label{subsec:comparison}
\begin{figure*}[!htp]
   \begin{center}
   {\includegraphics[width=0.245\linewidth]{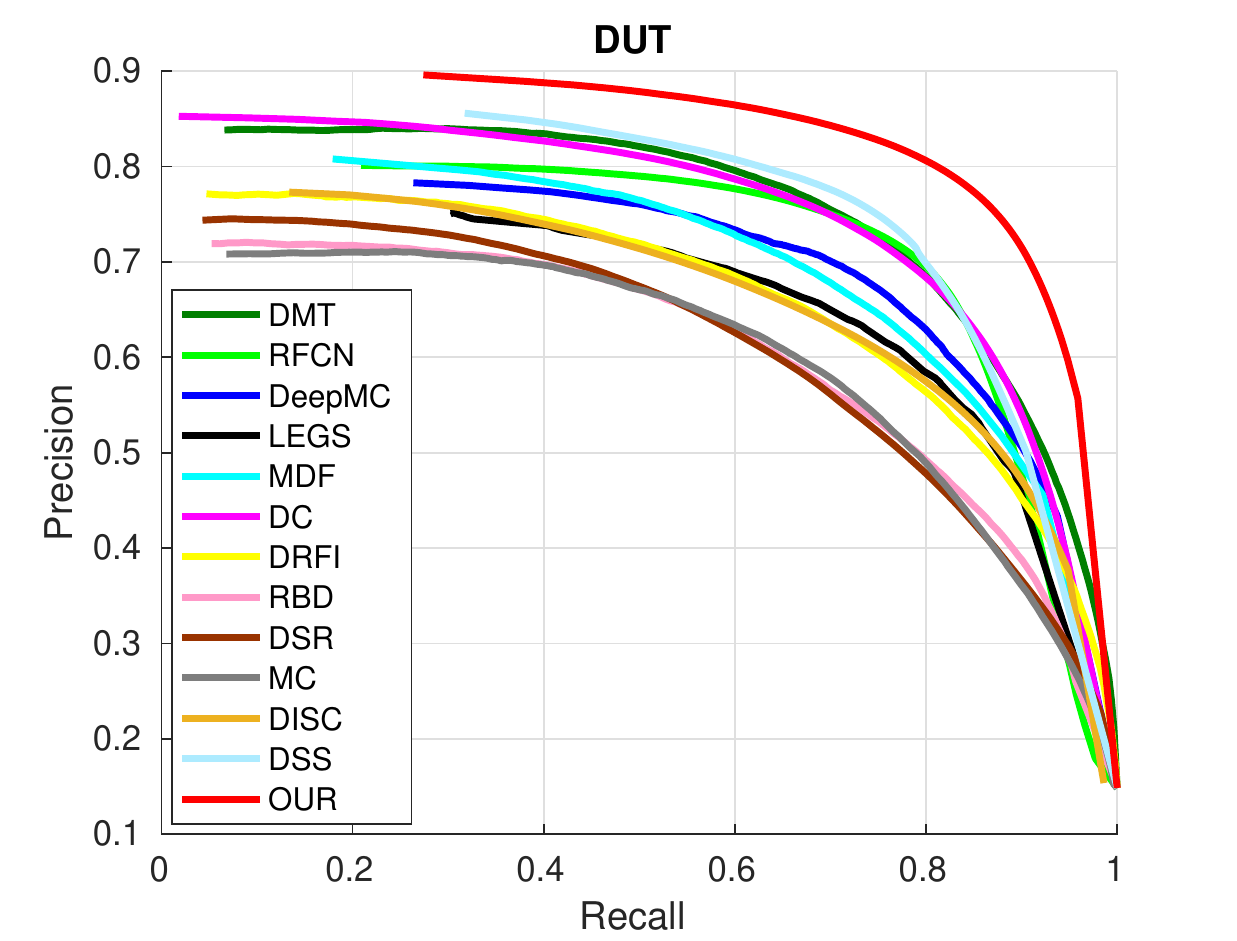}}
   {\includegraphics[width=0.245\linewidth]{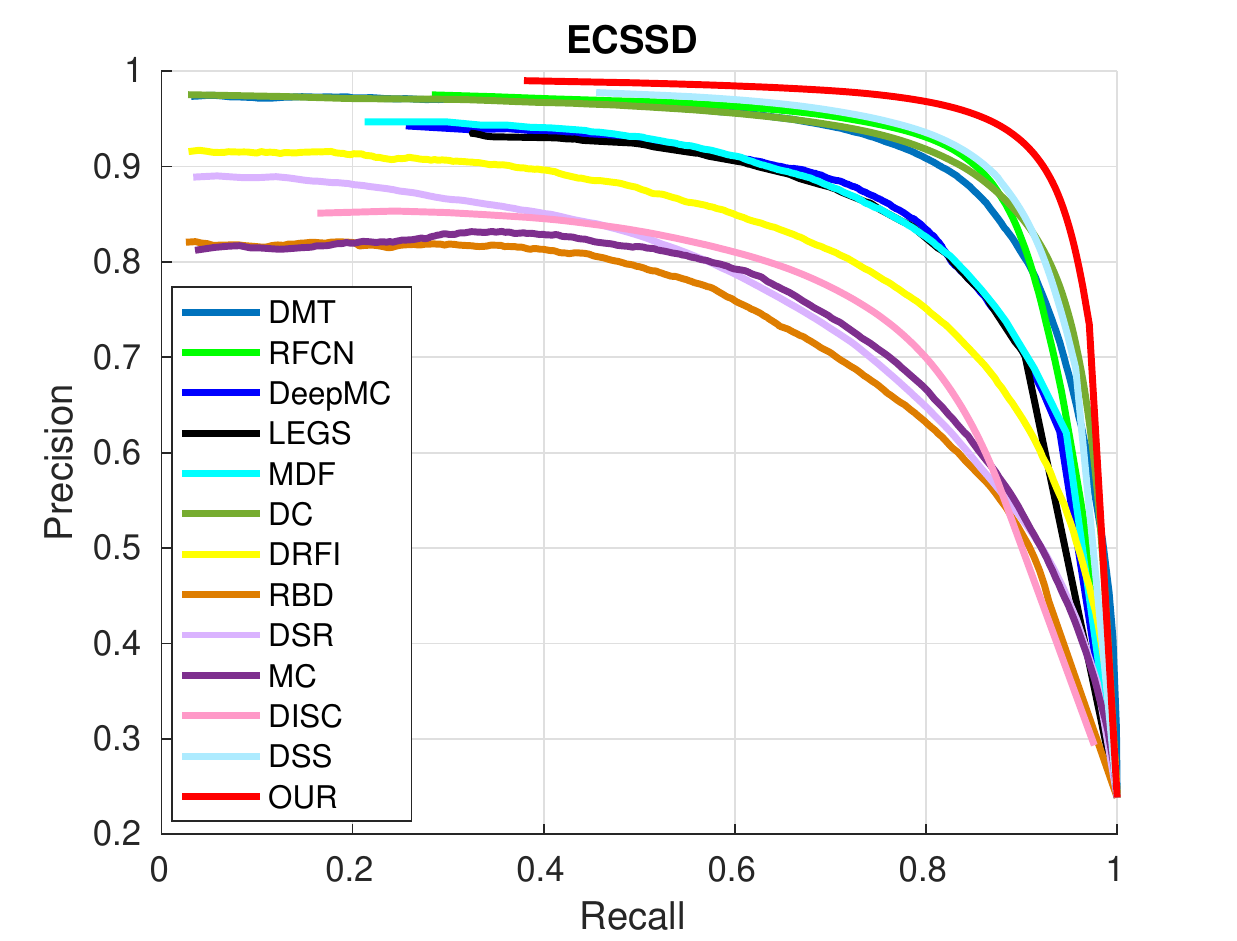}}
   {\includegraphics[width=0.245\linewidth]{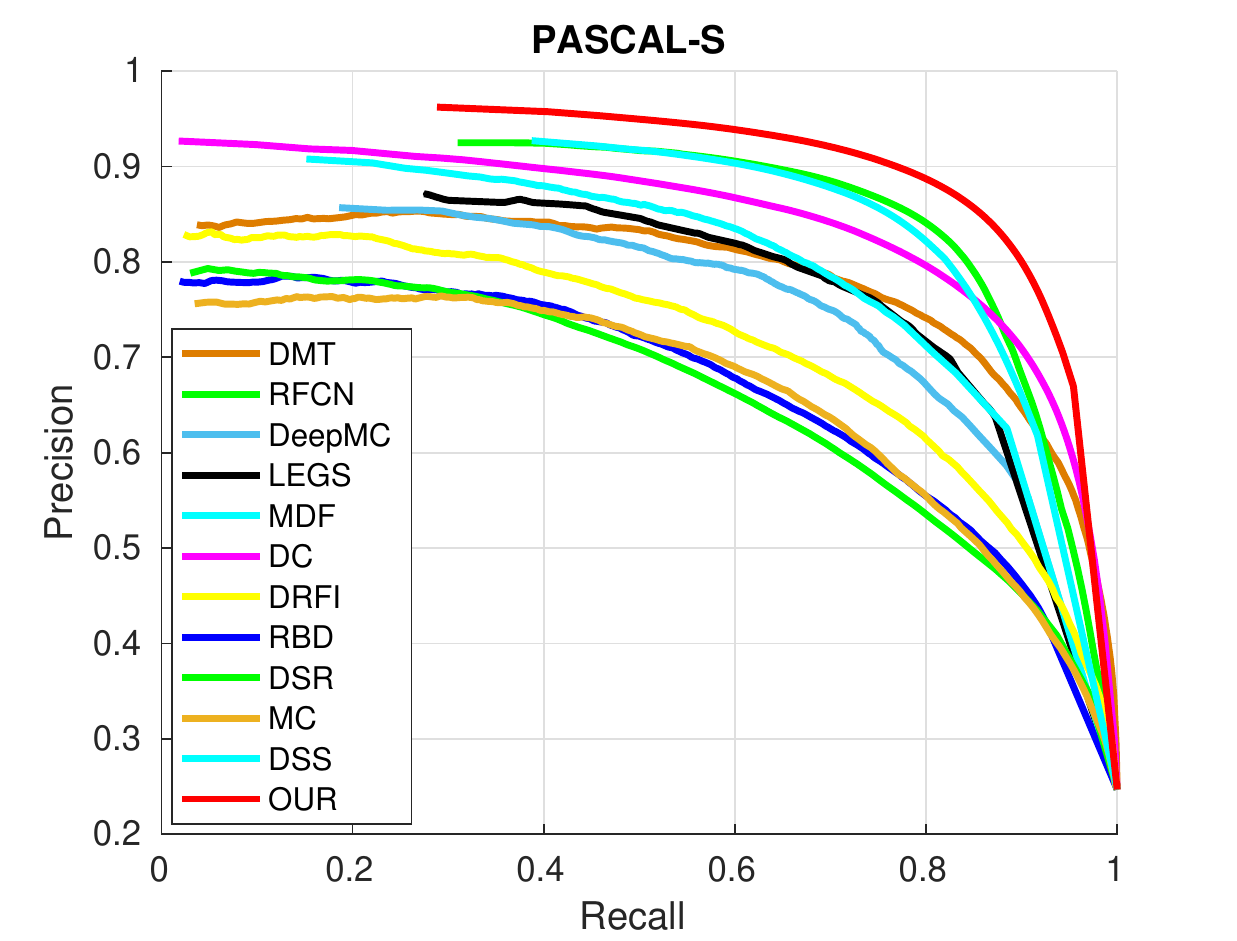}}
   {\includegraphics[width=0.245\linewidth]{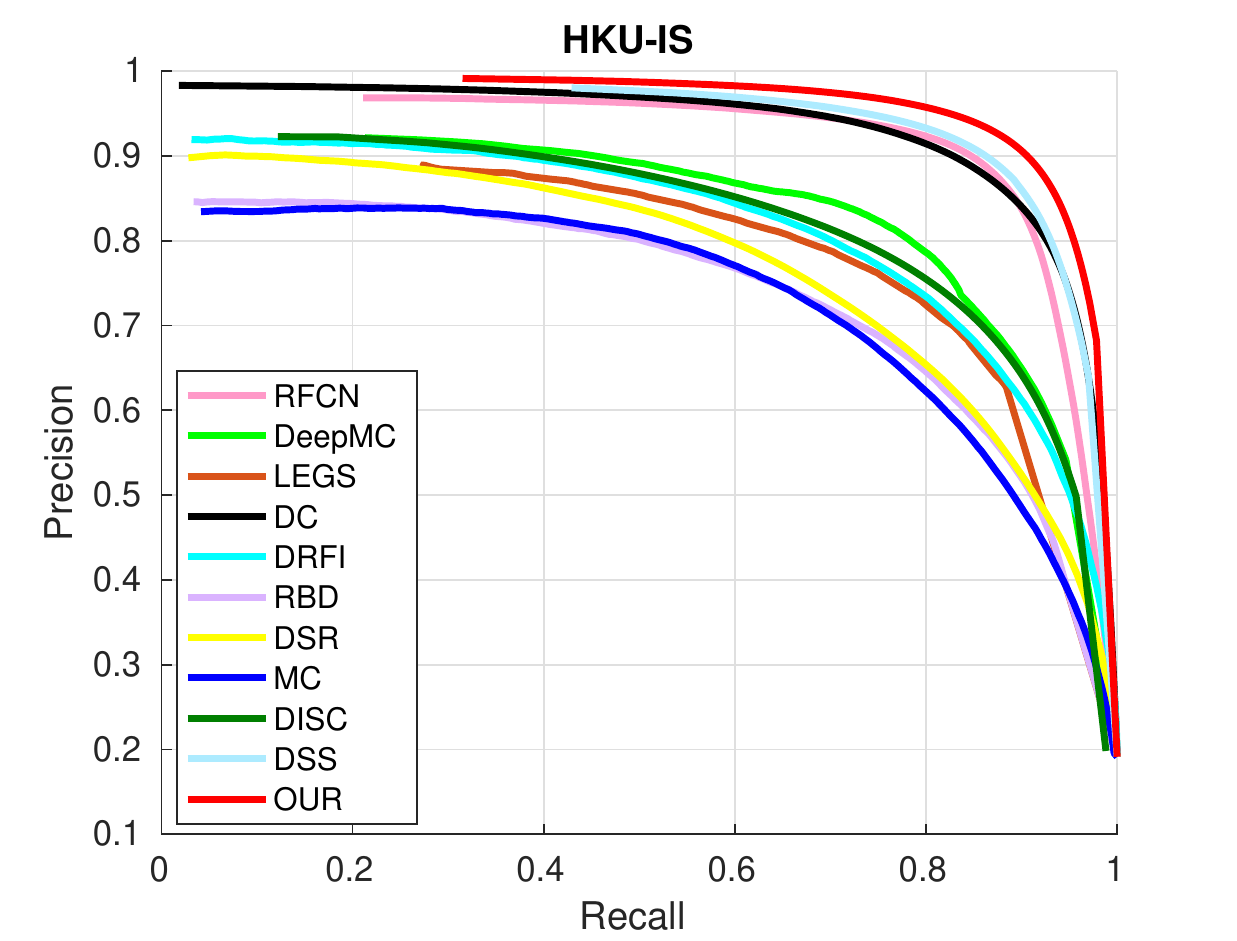}}
   {\includegraphics[width=0.245\linewidth]{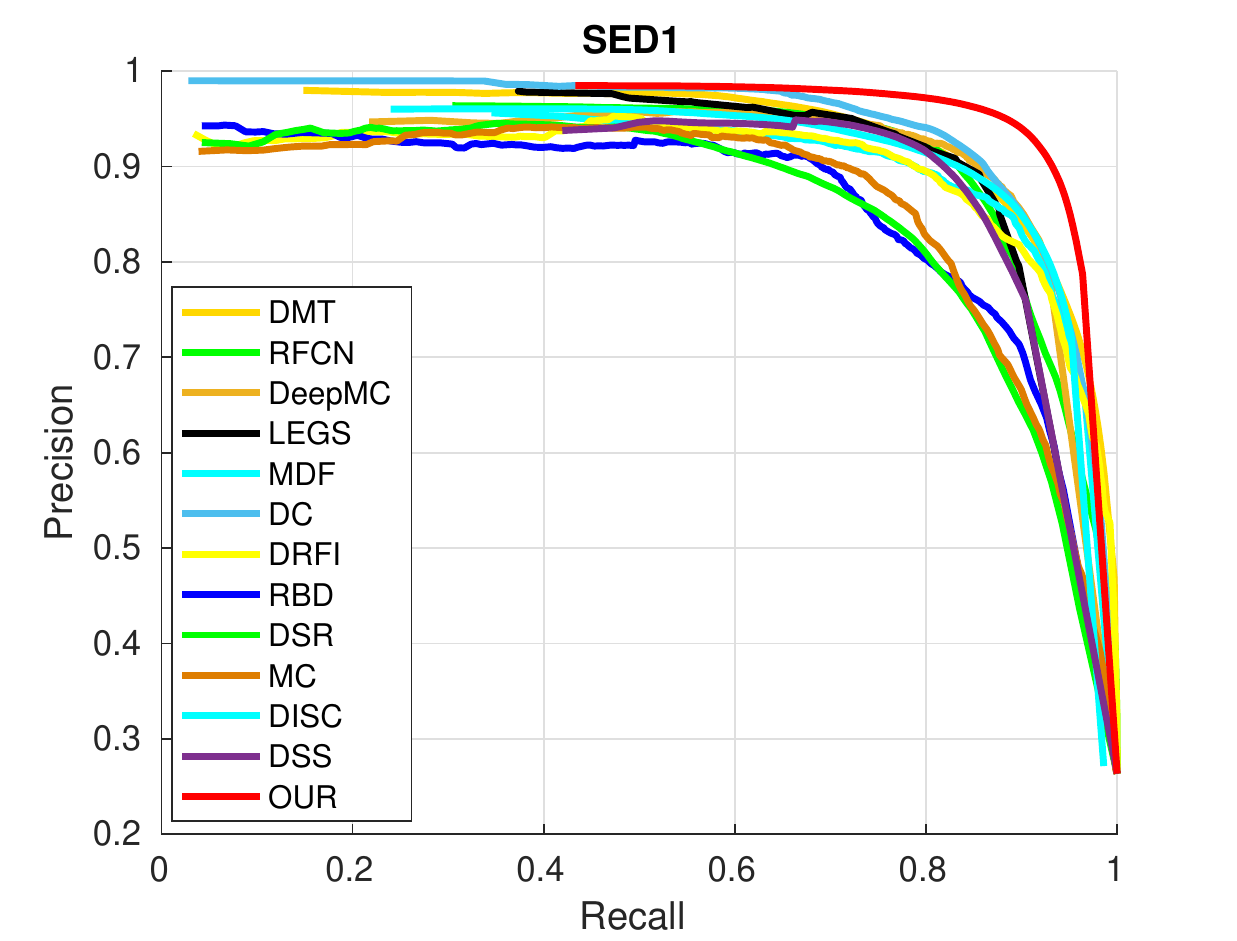}}
   {\includegraphics[width=0.245\linewidth]{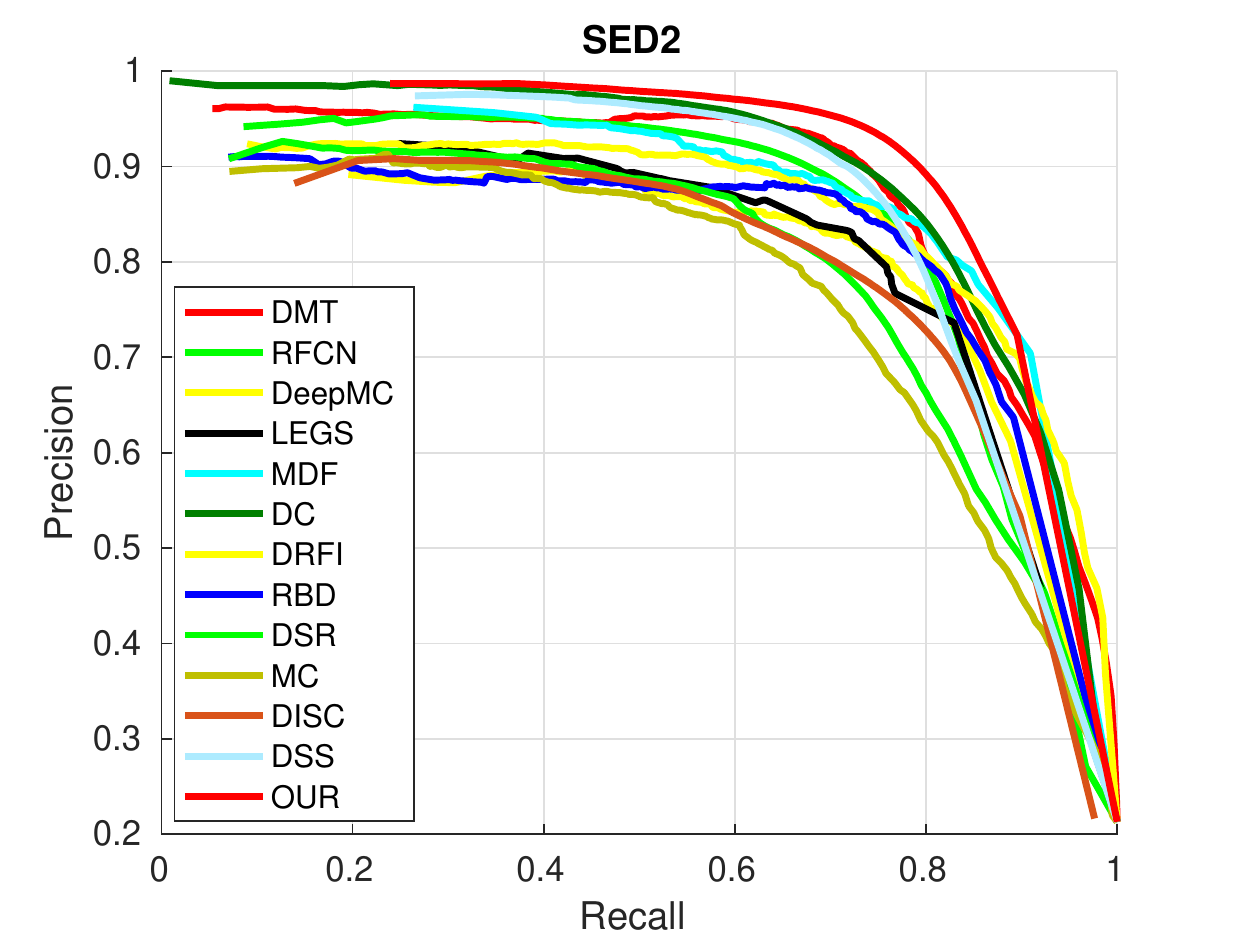}}
   {\includegraphics[width=0.245\linewidth]{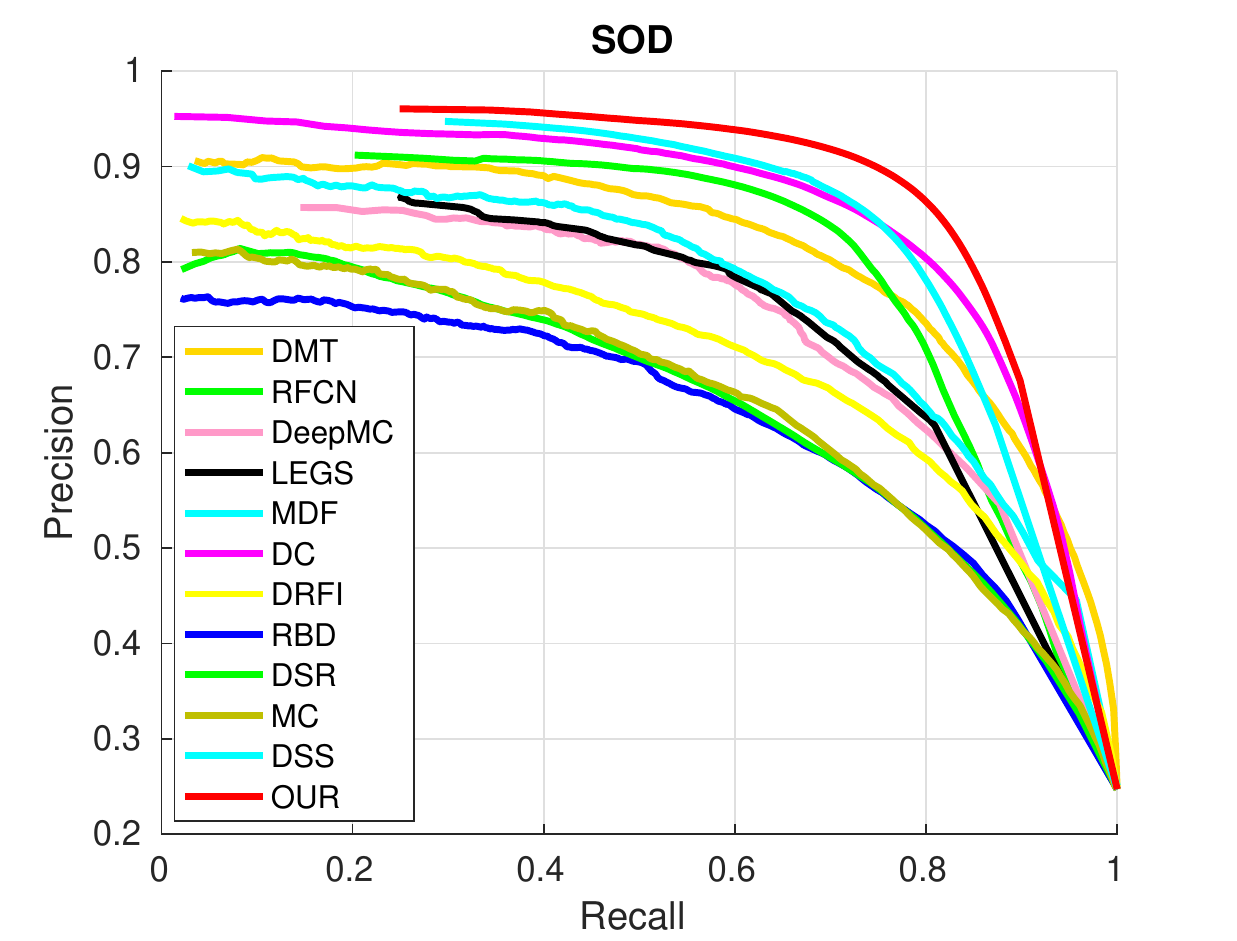}}
   {\includegraphics[width=0.245\linewidth]{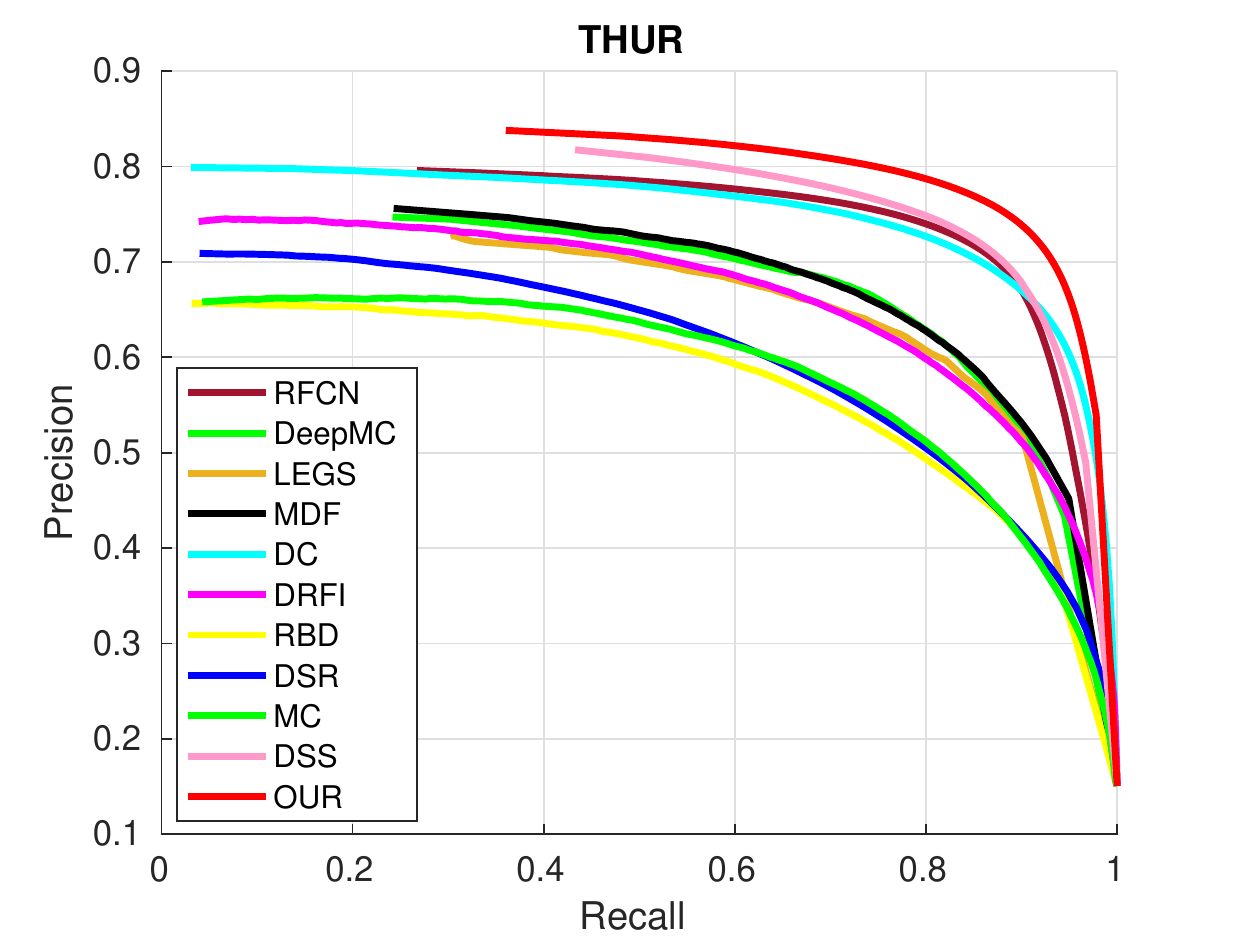}}
   \end{center}
   \vspace{-5mm}
   \caption{Precision-Recall curves on eight benchmark datasets (DUT, ECSSD, PASCAL-S, HKU-IS, ICOSEG, SED1, SOD, THUR). Our method consistently outperforms all the competing methods on all the datasets. Best Viewed on Screen.}
   \label{fig:pr_curve}
\end{figure*}

\textbf{Quantitative Comparison:} We compared our method with eleven deep saliency methods and five conventional methods. Results are reported in Table~\ref{tab:deep_unsuper_Performance_Comparison} and Fig.~\ref{fig:pr_curve}, where ``Ours'' represents the results of our model.

Table~\ref{tab:deep_unsuper_Performance_Comparison} shows that on those ten datasets, deep learning based methods outperform traditional methods with 3\%-9\% decrease in MAE, which further proves the superiority of deep features for saliency detection. DSS \cite{ChengCVPR17}, DC \cite{DC} and our model are all built on the DeepLab framework \cite{Deeplab}, while our method integrating handcrafted saliency with ResNet-101 model achieves the best results, especially on the THUR dataset, where our method achieves about 3\% performance leap of maximum F-measure and around 6\% improvement of mean F-measure, as well as around 3\% decrease in MAE compared with the above two DeepLab based methods. Also, our method consistently achieves the best mean F-measure compared with those state-of-the-art deep learning based methods as well as those handcrafted saliency methods. Fig.~\ref{fig:pr_curve} shows PR curves of our method and the competing methods on eight benchmark datasets, where our method consistently achieves best performance compared with the competing methods, especially on the DUT dataset, where our method outperforms the compared methods with a wide margin.
\begin{table*}[!htp] \small
\begin{center}
\caption{Performance for different methods including ours on ten benchmark datasets (Best ones in bold). Each cell: max F-measure (higher better) / mean F-measure (higher better) / MAE (lower better). } \centering
\begin{tabular}{l| c| c| c| c| c| c| c| c| c| c}\hline
  & MSRA-B & ECSSD & DUT & SED1 & SED2 & PASCAL-S & ICOSEG & HKU-IS & THUR & SOD \\ \hline
  &\textbf{0.9310} & \textbf{0.9233} & \textbf{0.8010}& \textbf{0.9237} & \textbf{0.8798}& \textbf{0.8671} & 0.8617 & \textbf{0.9187}& \textbf{0.7811}& \textbf{0.8562}\\
Ours & \textbf{0.9184} & \textbf{0.9082} & \textbf{0.7829}& \textbf{0.9085} & \textbf{0.8519}& \textbf{0.8504} & \textbf{0.8444} & \textbf{0.9002}& \textbf{0.7638}& \textbf{0.8367}\\
&\textbf{0.0379} & \textbf{0.0619} & \textbf{0.0600}& \textbf{0.0627}& \textbf{0.0861}& \textbf{0.1452} & 0.0751 & \textbf{0.0466}& \textbf{0.0673}& \textbf{0.1014}\\ \hline
  &0.9146  & 0.9006 & 0.7572 & 0.8879 & 0.8526 & 0.8418 & 0.8595 &0.8986 & 0.7357 & 0.8281\\
  DSS \cite{ChengCVPR17} &0.8941  & 0.8796 & 0.7290 & 0.8678 & 0.8236 & 0.8243 & 0.8322 &0.8718 & 0.7081 & 0.8048\\
  &0.0474  & 0.0699 & 0.0760 & 0.0887 & 0.1014 & 0.1546 & 0.0757 &0.0520 & 0.1142 & 0.1118\\ \hline
&0.8973 & 0.8095 & 0.7449 & - & 0.8634 & 0.8034 & - &- & 0.7276 & 0.7807\\
  DMT \cite{TIP}  &0.8364  & 0.7589 & 0.6045 & - & 0.7778 & 0.6657 &-  &- & 0.6254 & 0.6978\\
  &0.0658  & 0.1601 & 0.0758 & - & 0.1074 & 0.2103 & - & -& 0.0854 & 0.1503\\ \hline
  &-  & 0.8970 & 0.7379 & 0.8923 & 0.8364 & 0.8495 & 0.8432 & 0.8917 & 0.7538 & 0.8038\\
RFCN \cite{RFCN} & - &  0.8426& 0.6918& 0.8467 & 0.7616& 0.8064 & 0.8028 & 0.8277&0.7062 & 0.7531\\
 &-  & 0.0973 & 0.0945 & 0.1020 & 0.1140 &0.1662 & 0.0948 & 0.0798 & 0.1003 & 0.1394\\
\hline
  &0.9051 & 0.8085 & 0.6595 & 0.8852 & 0.7791 & - & 0.7963 & 0.7845& -& -\\
DISC \cite{DISC} &0.8644 & 0.7772& 0.6085 & 0.8659 & 0.7436 &-& 0.7576  & 0.7359 &- & -\\
&0.0536  & 0.1137 & 0.1187 & 0.0777 & 0.1208 &  -& 0.1159 & 0.1030 &- & -\\
\hline
&0.9229  & 0.8379 & 0.7028 & 0.8967 & 0.7991 & 0.7605 & 0.7946 & 0.8080 &0.6855 & 0.7277\\
DeepMC \cite{DeepMC} & 0.8966 & 0.8061 & 0.6715 & 0.8600 & 0.7660 & 0.7327 & 0.7648 & 0.7676 & 0.6549 & 0.6862\\
&0.0491 & 0.1019 & 0.0885 & 0.0881 & 0.1162 & 0.1928 & 0.1049 & 0.0913 & 0.1025 & 0.1557\\
\hline
&0.8712  & 0.8303 &0.6677 &  0.8897& 0.8031 & 0.7760 & 0.7571 & 0.7662& 0.6638& 0.7347\\
LEGS \cite{LEGS} &0.8258  & 0.7855 & 0.6265& 0.8453 & 0.7357 & 0.7215 &  0.7093&0.7188 & 0.6301& 0.6870\\
&0.0800 & 0.1187 &0.1318 & 0.0997 & 0.1251 & 0.2005 & 0.1269 & 0.1186& 0.1242& 0.1729\\
\hline

&0.8853  & 0.8307 &0.6944 & 0.8916 & 0.8432 & 0.7900 & 0.8376 & - & 0.6847& 0.7381\\
MDF \cite{MDF:CVPR-2015} &0.7780  & 0.8097 &0.6768 & 0.7888 & 0.7658 & 0.7425 & 0.7847 & -& 0.6670& 0.6377\\
&0.1040  & 0.1081 &0.0916 & 0.1198 & 0.1171 & 0.2069 & 0.1008 & -& 0.1029& 0.1669\\
\hline
& 0.9045 & 0.8796 & -&  -& 0.8541& - &  -&0.8564 & 0.7160&-\\
RACDN \cite{RACDNN} & 0.8997  & 0.8755 & -&  -&0.8465 & - &-  & 0.8516& 0.7096&- \\
& 0.0514 & 0.0713 & -&-  & 0.0868& - & - & 0.0636 & 0.0866&-\\
\hline

&-  & 0.8674 &0.7195 &-  &- & 0.7898 & - &- & 0.7312&-\\
ELD \cite{ELD} & - & 0.8372 & 0.6651& - &- & 0.7784 &-  & -&0.6805 & -\\
& - &0.0805 &0.0909 & - & -& 0.1690 & - & -& 0.0952&-\\
\hline

& 0.9176& 0.8879& 0.7391& 0.9045 & 0.8567&0.8360 & \textbf{0.8727} & 0.8853& 0.7441&0.8219\\
DC \cite{DC} &0.8973  & 0.8315 &0.6902 & 0.8564 &0.7840 & 0.7861 & 0.8291 & 0.8205&0.6940 & 0.7603\\
 & 0.0467& 0.0906& 0.0971& 0.0886 &0.1014 &0.1614 & \textbf{0.0740} & 0.0730&0.0959 &0.1208\\
\hline
&0.8514 & 0.7834 &0.6638 & 0.8731 &0.8265 & 0.7306 & 0.8108 &0.7771 & 0.6657&0.6823 \\
DRFI \cite{DRFI:CVPR-2013} & 0.7282 & 0.6440 &0.5525 & 0.7397 &0.7252 & 0.5745 & 0.6986 & 0.6397& 0.5613& 0.5440\\
&0.1229  & 0.1719 &0.1496 & 0.1454 & 0.1373& 0.2556 & 0.1397 &0.1445 & 0.1471&0.2046 \\ \hline
&- & 0.7397 &0.6207 & 0.8253 &0.8009 & 0.6981 & 0.7854 &0.7405 & 0.6233&0.6577 \\
HDCT \cite{High-Dim-Color-Transform:CVPR-2014} & - & 0.5926 &0.5092 & 0.6844 &0.6668 & 0.5380 & 0.6669 & 0.5931& 0.5109& 0.5210\\
&- & 0.1998 &0.1666 & 0.1761 & 0.1638& 0.2761 & 0.1611 &0.1648 & 0.1670&0.2230 \\
\hline
&0.8034  & 0.6961 & 0.6031 & 0.8200 & 0.8201& 0.6856 & 0.7843 & 0.7074&0.5813 & 0.6215\\
RBD \cite{Background-Detection:CVPR-2014} & 0.7508 & 0.6518 & 0.5100& 0.7747 & 0.7939& 0.6581 & 0.7440 & 0.6445& 0.5221& 0.5927\\
&0.1171 & 0.1832 & 0.2011& 0.1316 & 0.1096& 0.2418 & 0.1189 & 0.1597& 0.1936& 0.2181\\ \hline
&0.8125 & 0.7345 &0.6261 & 0.8299 &0.7852 & 0.6906 & 0.7658 &0.7414 & 0.6125&0.6440 \\
DSR \cite{DSR_ICCV13} & 0.7337 & 0.6387 &0.5583 & 0.7277 &0.7053 & 0.5785 & 0.7002 & 0.6438& 0.5498& 0.5500\\
&0.1207 & 0.1742 &0.1374 & 0.1614 & 0.1457& 0.2600 & 0.1491 &0.1404 & 0.1408&0.2133 \\
\hline
&0.8264& 0.7416 &0.6273 &0.8502 & 0.7699 & 0.7097 &0.7857 & 0.7234&0.6096& 0.6493  \\
MC \cite{MC_ICCV13} &0.7165  & 0.6114 &0.5289 & 0.7319 &0.6619 & 0.5742 & 0.6790 &0.5900 & 0.5149& 0.5332\\
&0.1441  & 0.2037 &0.1863 & 0.1620 & 0.1848& 0.2719 & 0.1729 & 0.1840&0.1838 & 0.2435 \\ \hline
\end{tabular}
\label{tab:deep_unsuper_Performance_Comparison}
\end{center}
\end{table*}

\begin{figure*}[!htp] \small
\begin{center}
\begin{tabular}{c c c c c c c c c c}
 \rotatebox{90}{Image} & \hspace{-0.4cm} \includegraphics[height=0.08\linewidth]{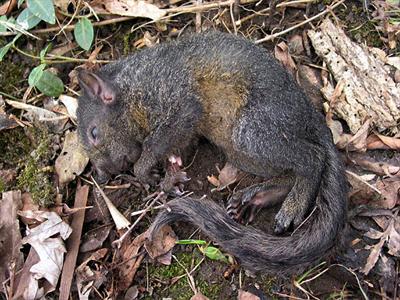}  &\hspace{-0.4cm} \includegraphics[height=0.080 \linewidth]{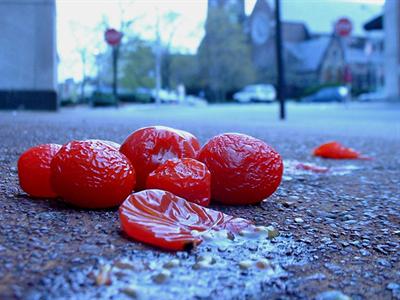}&\hspace{-0.4cm} \includegraphics[height=0.080\linewidth]{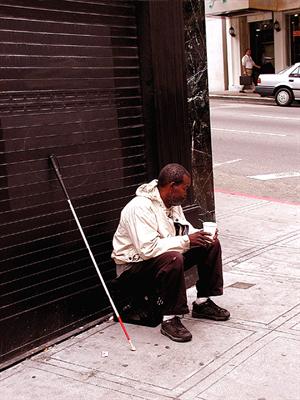} & \hspace{-0.38cm} \includegraphics[height=0.080\linewidth]
{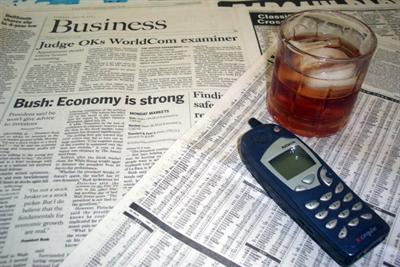}&\hspace{-0.28cm}\includegraphics[height=0.080\linewidth]{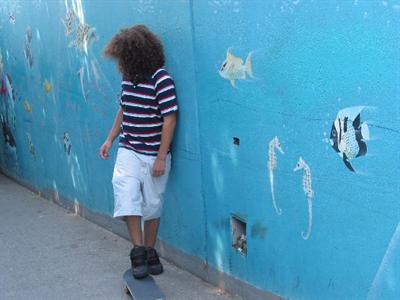}&\hspace{-0.38cm}  \includegraphics[height=0.080\linewidth]{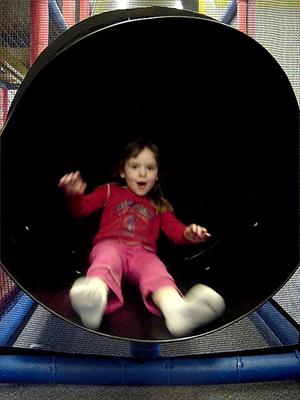}
  &\hspace{-0.4cm} \includegraphics[height=0.080\linewidth]{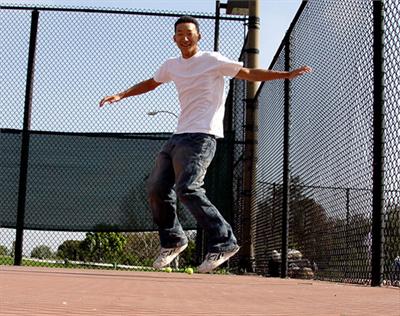}
  &\hspace{-0.4cm}    \includegraphics[height=0.080\linewidth]{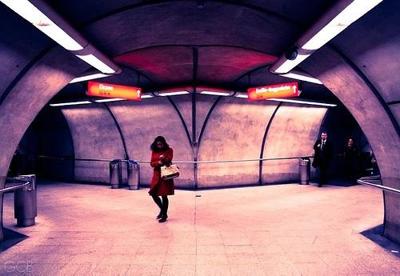}
   & \hspace{-0.4cm}    \includegraphics[height=0.080\linewidth]{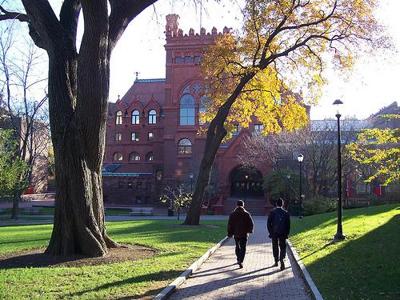}\\
    \rotatebox{90}{GT} & \hspace{-0.4cm} \includegraphics[height=0.08\linewidth]{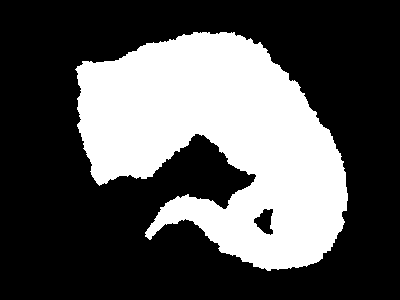}  &\hspace{-0.4cm} \includegraphics[height=0.080 \linewidth]{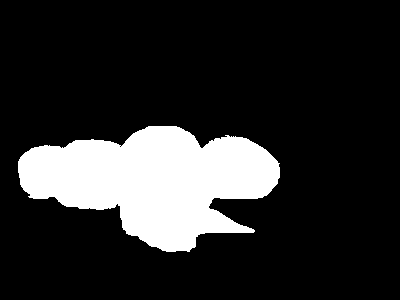}&\hspace{-0.4cm} \includegraphics[height=0.080\linewidth]{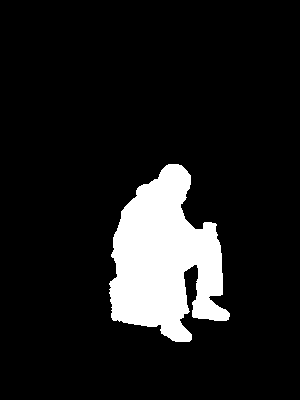} & \hspace{-0.38cm} \includegraphics[height=0.080\linewidth]
{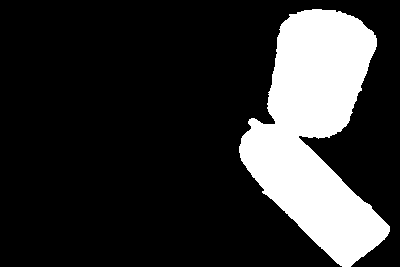}&\hspace{-0.28cm}\includegraphics[height=0.080\linewidth]{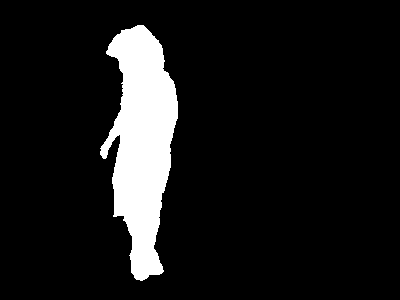}&\hspace{-0.38cm}  \includegraphics[height=0.080\linewidth]{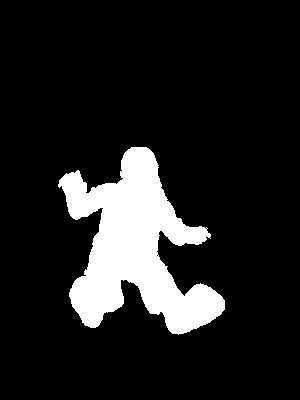}
  &\hspace{-0.4cm} \includegraphics[height=0.080\linewidth]{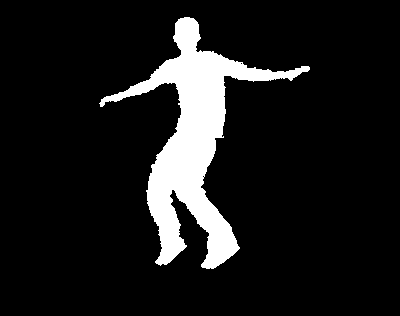}
  &\hspace{-0.4cm}    \includegraphics[height=0.080\linewidth]{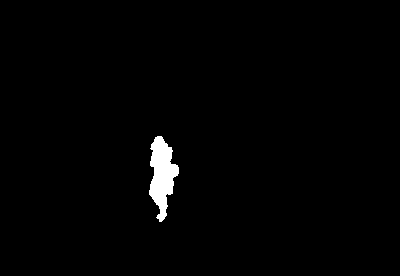}
   & \hspace{-0.4cm}    \includegraphics[height=0.080\linewidth]{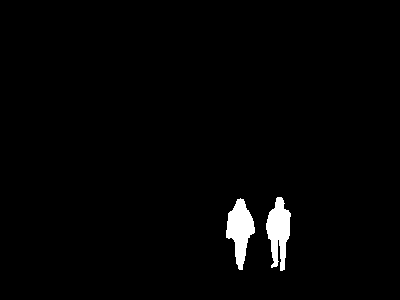}\\
                          \rotatebox{90}{OUR} & \hspace{-0.4cm} \includegraphics[height=0.08\linewidth]{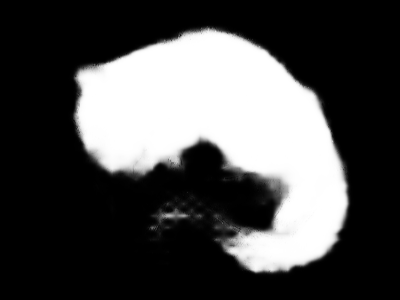}  &\hspace{-0.4cm} \includegraphics[height=0.080 \linewidth]{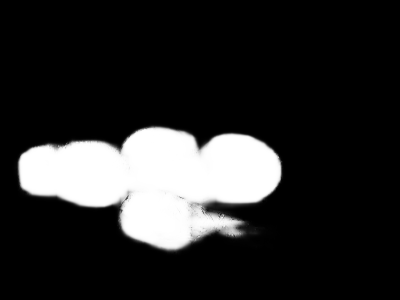}&\hspace{-0.4cm} \includegraphics[height=0.080\linewidth]{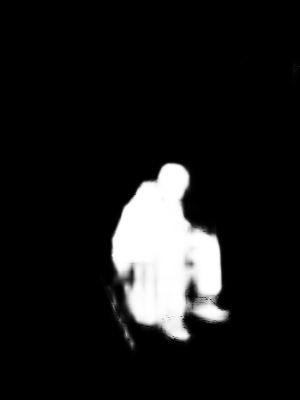} & \hspace{-0.38cm} \includegraphics[height=0.080\linewidth]
{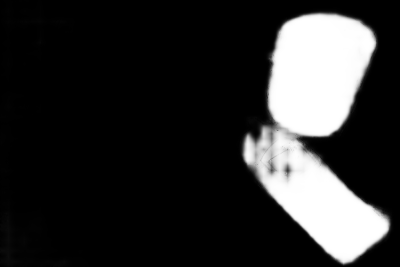}&\hspace{-0.28cm}\includegraphics[height=0.080\linewidth]{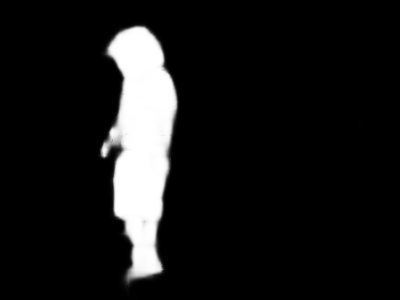}&\hspace{-0.38cm}  \includegraphics[height=0.080\linewidth]{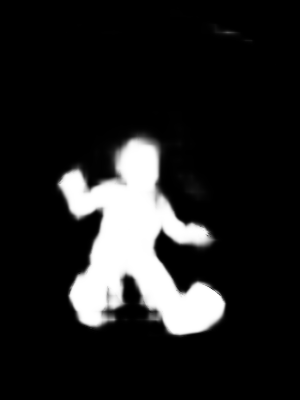}
  &\hspace{-0.4cm} \includegraphics[height=0.080\linewidth]{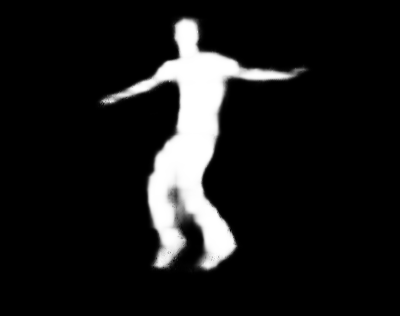}
  &\hspace{-0.4cm}    \includegraphics[height=0.080\linewidth]{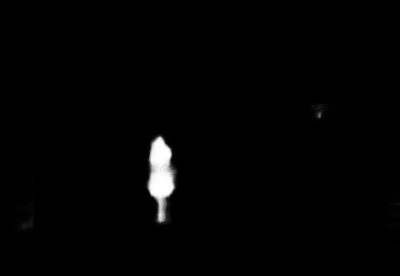}
   & \hspace{-0.4cm}    \includegraphics[height=0.080\linewidth]{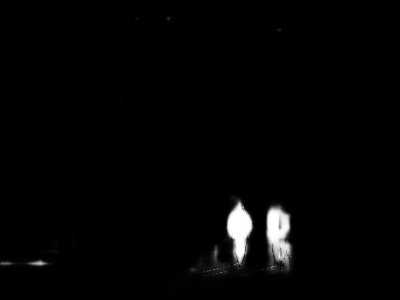}\\
       \rotatebox{90}{DISC} & \hspace{-0.4cm} \includegraphics[height=0.08\linewidth]{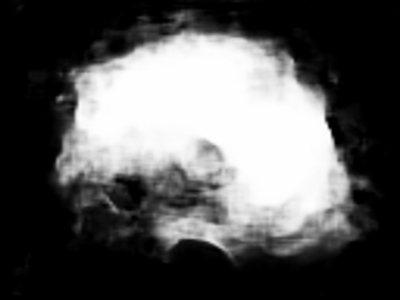}  &\hspace{-0.4cm} \includegraphics[height=0.080 \linewidth]{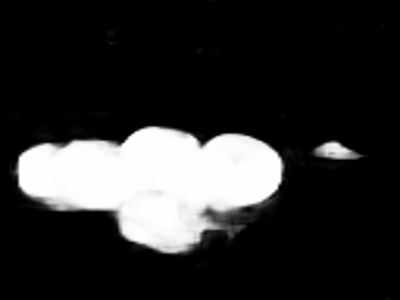}&\hspace{-0.4cm} \includegraphics[height=0.080\linewidth]{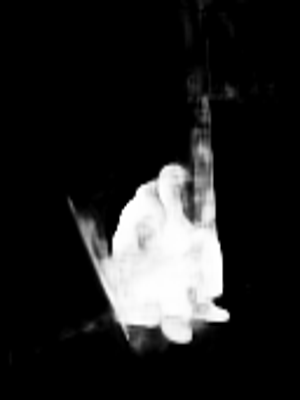} & \hspace{-0.38cm} \includegraphics[height=0.080\linewidth]
{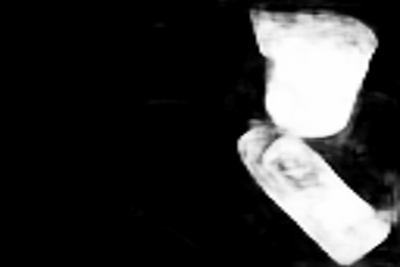}&\hspace{-0.28cm}\includegraphics[height=0.080\linewidth]{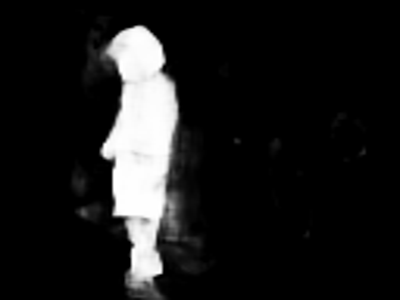}&\hspace{-0.38cm}  \includegraphics[height=0.080\linewidth]{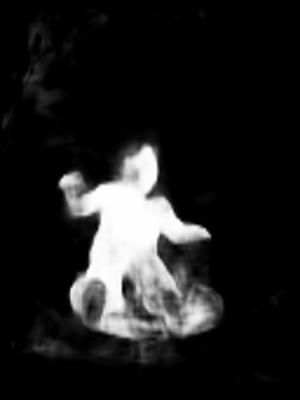}
  &\hspace{-0.4cm} \includegraphics[height=0.080\linewidth]{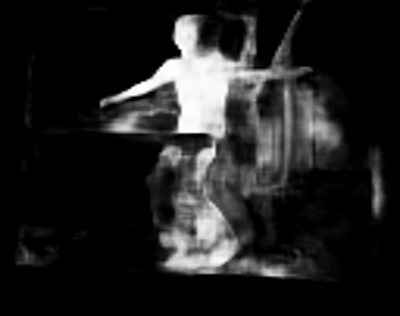}
  &\hspace{-0.4cm}    \includegraphics[height=0.080\linewidth]{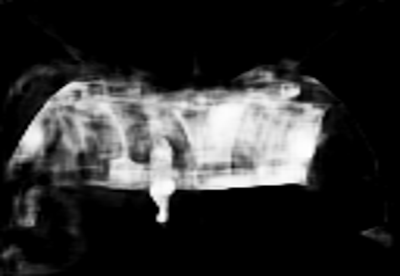}
   & \hspace{-0.4cm}    \includegraphics[height=0.080\linewidth]{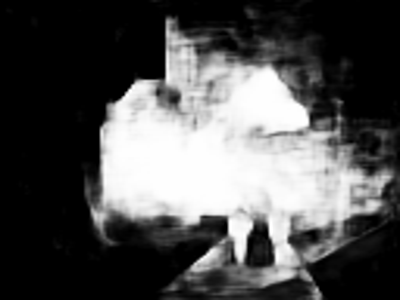}\\
          \rotatebox{90}{LEGS} & \hspace{-0.4cm} \includegraphics[height=0.08\linewidth]{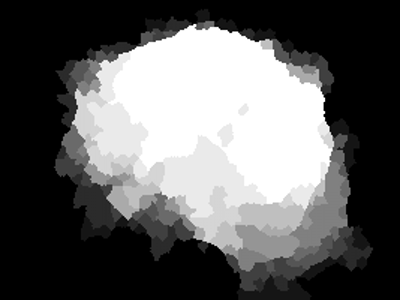}  &\hspace{-0.4cm} \includegraphics[height=0.080 \linewidth]{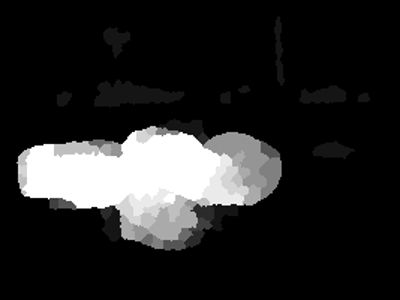}&\hspace{-0.4cm} \includegraphics[height=0.080\linewidth]{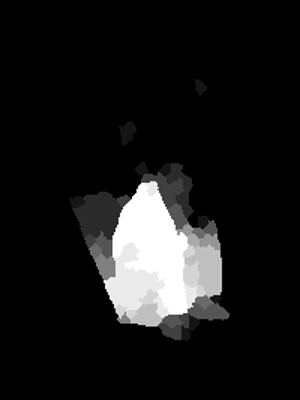} & \hspace{-0.38cm} \includegraphics[height=0.080\linewidth]
{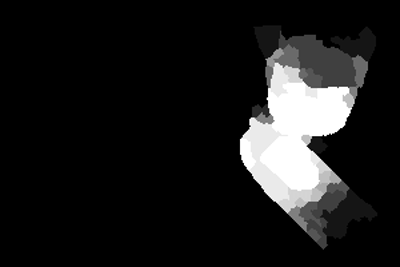}&\hspace{-0.28cm}\includegraphics[height=0.080\linewidth]{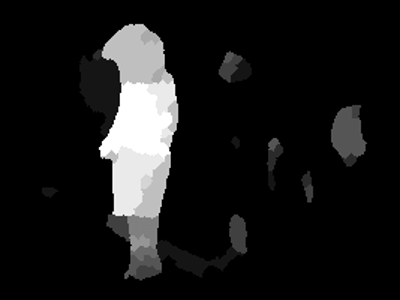}&\hspace{-0.38cm}  \includegraphics[height=0.080\linewidth]{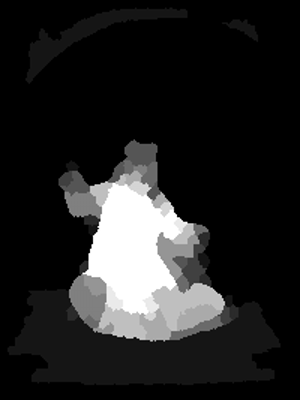}
  &\hspace{-0.4cm} \includegraphics[height=0.080\linewidth]{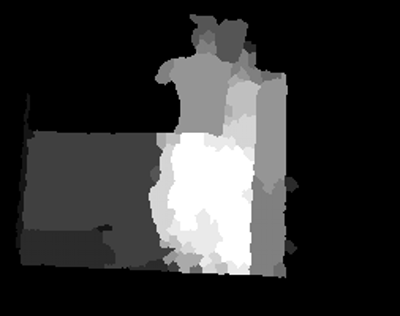}
  &\hspace{-0.4cm}    \includegraphics[height=0.080\linewidth]{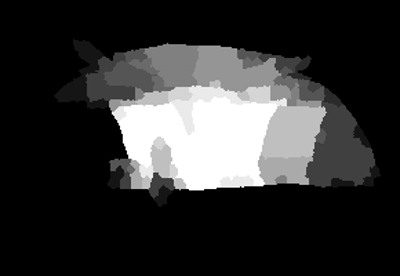}
   & \hspace{-0.4cm}    \includegraphics[height=0.080\linewidth]{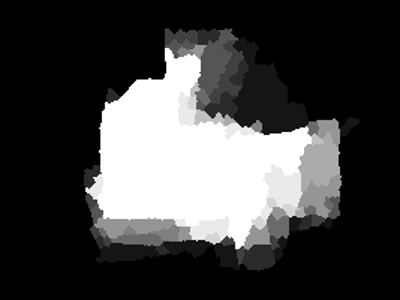}\\
            \rotatebox{90}{MDF} & \hspace{-0.4cm} \includegraphics[height=0.08\linewidth]{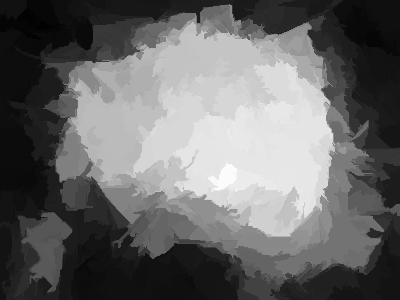}  &\hspace{-0.4cm} \includegraphics[height=0.080 \linewidth]{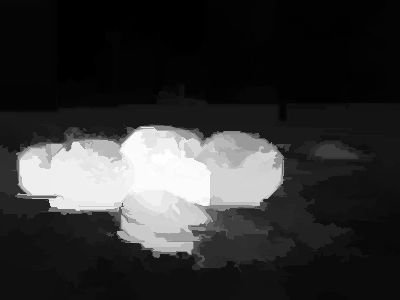}&\hspace{-0.4cm} \includegraphics[height=0.080\linewidth]{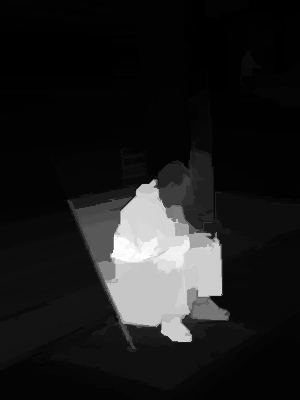} & \hspace{-0.38cm} \includegraphics[height=0.080\linewidth]
{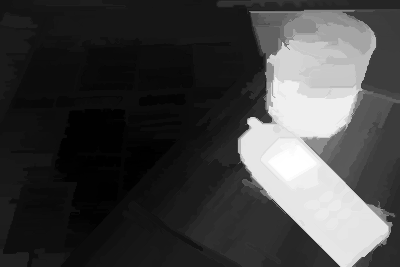}&\hspace{-0.28cm}\includegraphics[height=0.080\linewidth]{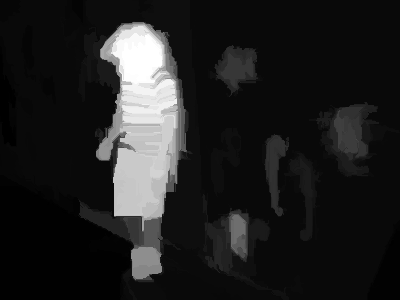}&\hspace{-0.38cm}  \includegraphics[height=0.080\linewidth]{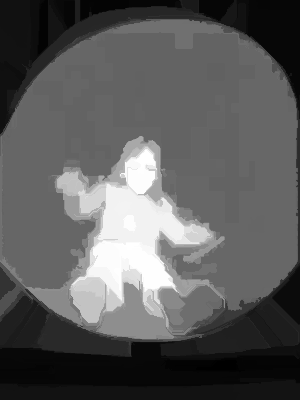}
  &\hspace{-0.4cm} \includegraphics[height=0.080\linewidth]{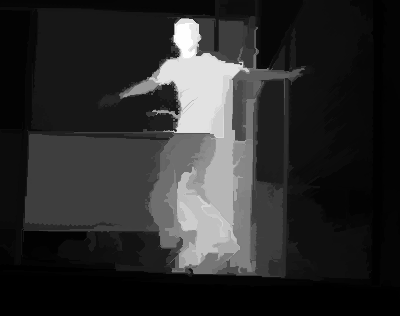}
  &\hspace{-0.4cm}    \includegraphics[height=0.080\linewidth]{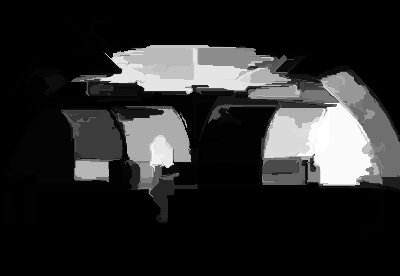}
   & \hspace{-0.4cm}    \includegraphics[height=0.080\linewidth]{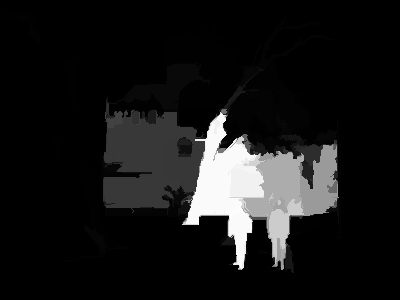}\\
            \rotatebox{90}{RFCN} & \hspace{-0.4cm} \includegraphics[height=0.08\linewidth]{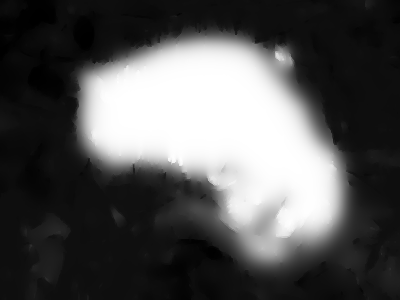}  &\hspace{-0.4cm} \includegraphics[height=0.080 \linewidth]{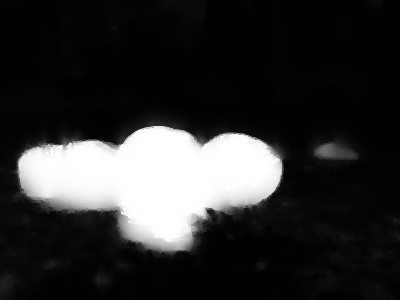}&\hspace{-0.4cm} \includegraphics[height=0.080\linewidth]{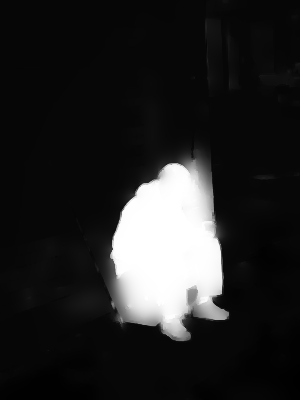} & \hspace{-0.38cm} \includegraphics[height=0.080\linewidth]
{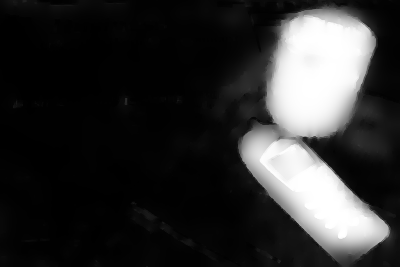}&\hspace{-0.28cm}\includegraphics[height=0.080\linewidth]{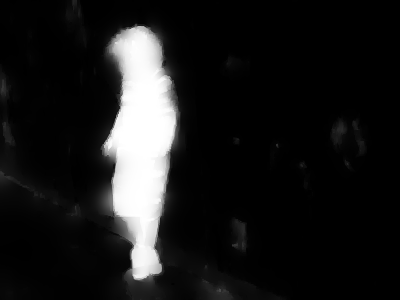}&\hspace{-0.38cm}  \includegraphics[height=0.080\linewidth]{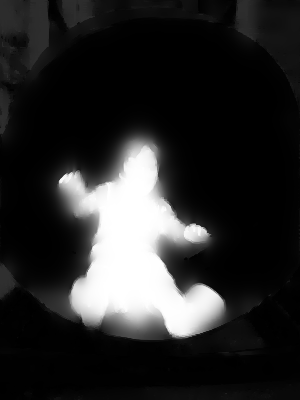}
  &\hspace{-0.4cm} \includegraphics[height=0.080\linewidth]{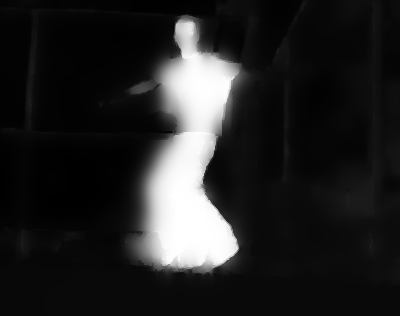}
  &\hspace{-0.4cm}    \includegraphics[height=0.080\linewidth]{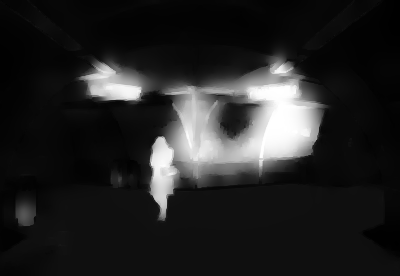}
   & \hspace{-0.4cm}    \includegraphics[height=0.080\linewidth]{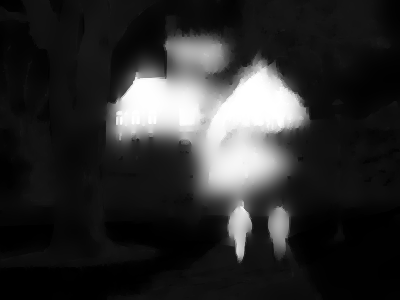}\\
            \rotatebox{90}{DMT} & \hspace{-0.4cm} \includegraphics[height=0.08\linewidth]{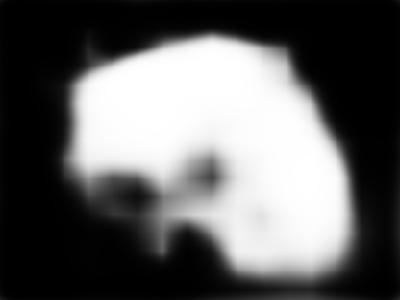}  &\hspace{-0.4cm} \includegraphics[height=0.080 \linewidth]{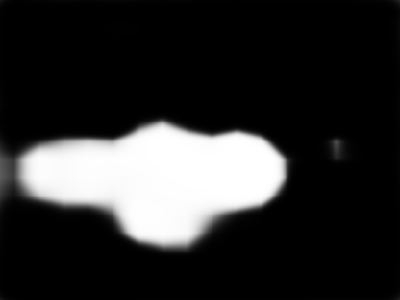}&\hspace{-0.4cm} \includegraphics[height=0.080\linewidth]{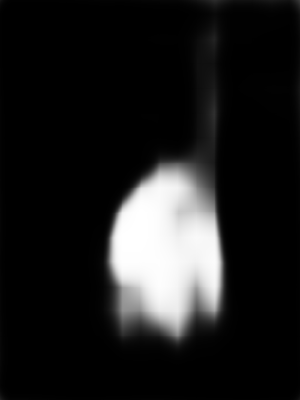} & \hspace{-0.38cm} \includegraphics[height=0.080\linewidth]
{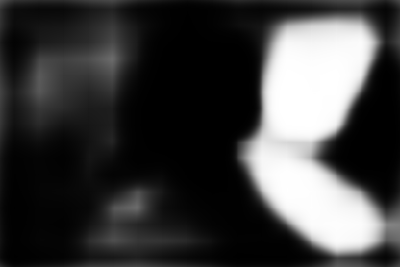}&\hspace{-0.28cm}\includegraphics[height=0.080\linewidth]{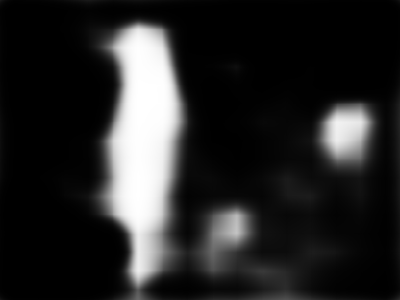}&\hspace{-0.38cm}  \includegraphics[height=0.080\linewidth]{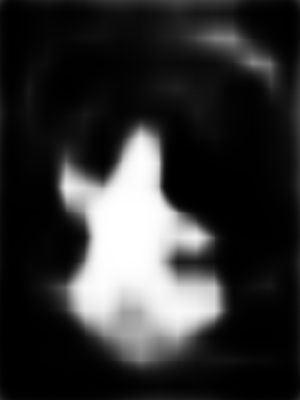}
  &\hspace{-0.4cm} \includegraphics[height=0.080\linewidth]{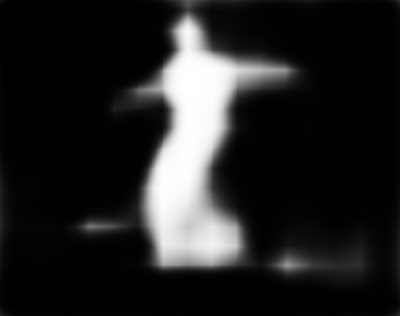}
  &\hspace{-0.4cm}    \includegraphics[height=0.080\linewidth]{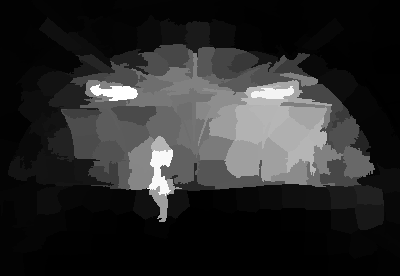}
   & \hspace{-0.4cm}    \includegraphics[height=0.080\linewidth]{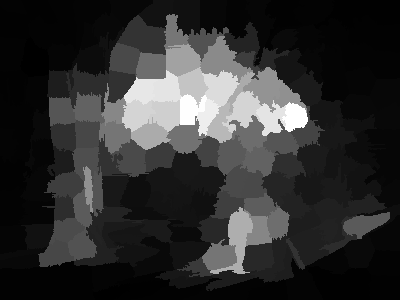}\\
             \rotatebox{90}{DC} & \hspace{-0.4cm} \includegraphics[height=0.08\linewidth]{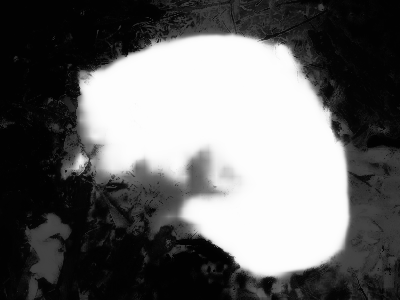}  &\hspace{-0.4cm} \includegraphics[height=0.080 \linewidth]{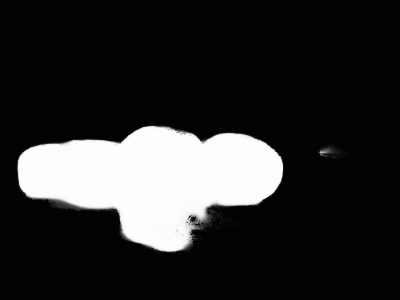}&\hspace{-0.4cm} \includegraphics[height=0.080\linewidth]{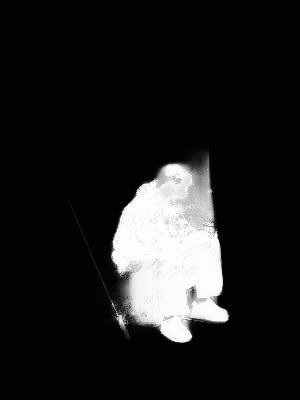} & \hspace{-0.38cm} \includegraphics[height=0.080\linewidth]
{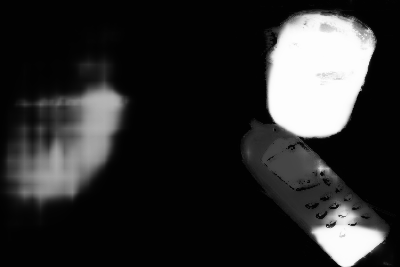}&\hspace{-0.28cm}\includegraphics[height=0.080\linewidth]{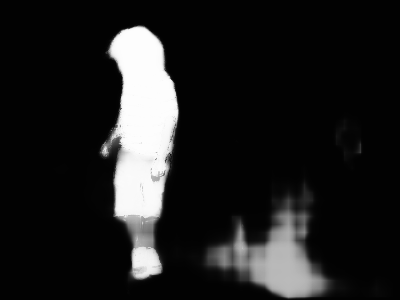}&\hspace{-0.38cm}  \includegraphics[height=0.080\linewidth]{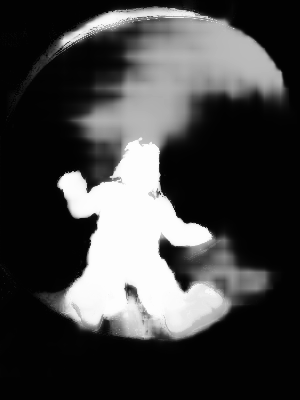}
  &\hspace{-0.4cm} \includegraphics[height=0.080\linewidth]{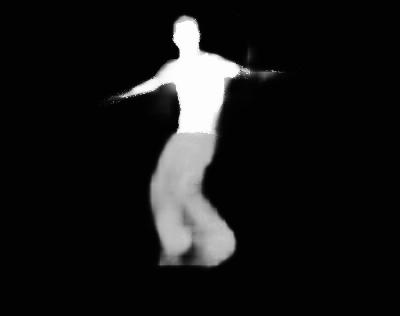}
  &\hspace{-0.4cm}    \includegraphics[height=0.080\linewidth]{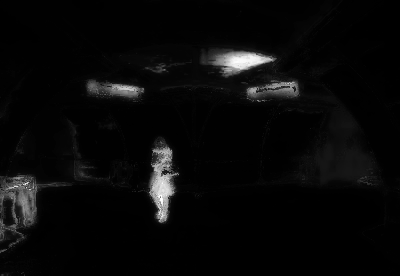}
   & \hspace{-0.4cm}    \includegraphics[height=0.080\linewidth]{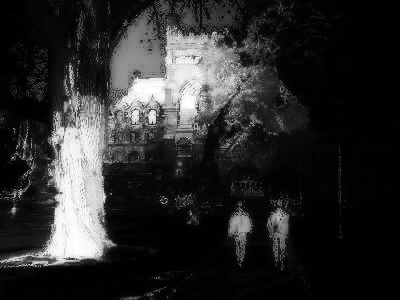}\\
                \rotatebox{90}{DSS} & \hspace{-0.4cm} \includegraphics[height=0.08\linewidth]{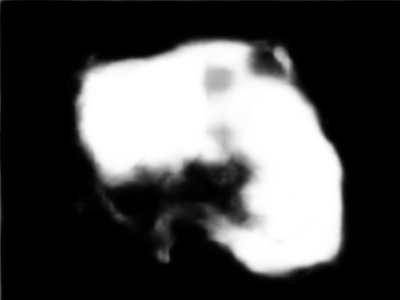}  &\hspace{-0.4cm} \includegraphics[height=0.080 \linewidth]{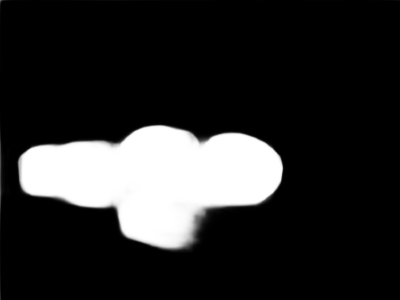}&\hspace{-0.4cm} \includegraphics[height=0.080\linewidth]{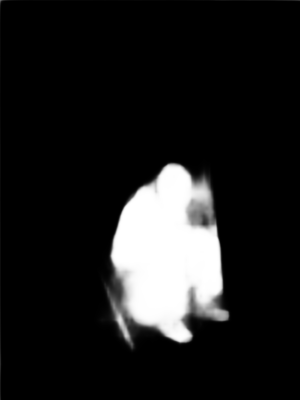} & \hspace{-0.38cm} \includegraphics[height=0.080\linewidth]
{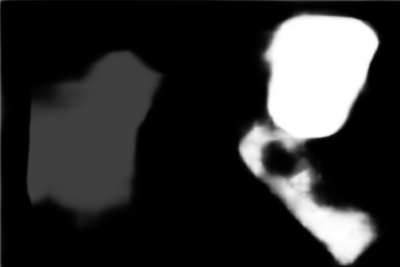}&\hspace{-0.28cm}\includegraphics[height=0.080\linewidth]{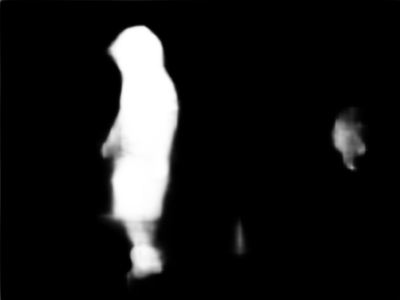}&\hspace{-0.38cm}  \includegraphics[height=0.080\linewidth]{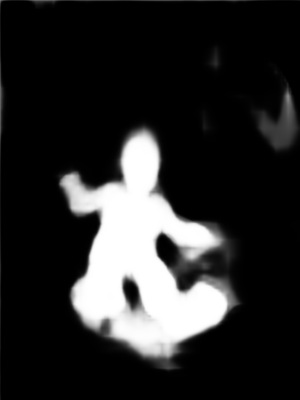}
  &\hspace{-0.4cm} \includegraphics[height=0.080\linewidth]{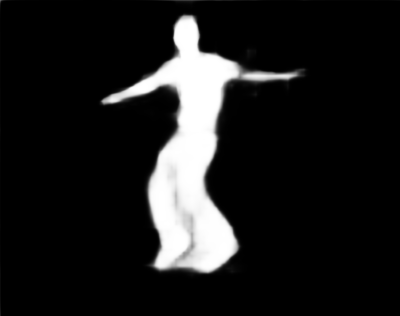}
  &\hspace{-0.4cm}    \includegraphics[height=0.080\linewidth]{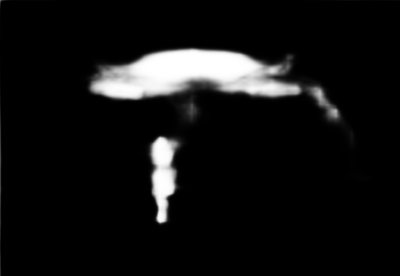}
   & \hspace{-0.4cm}    \includegraphics[height=0.080\linewidth]{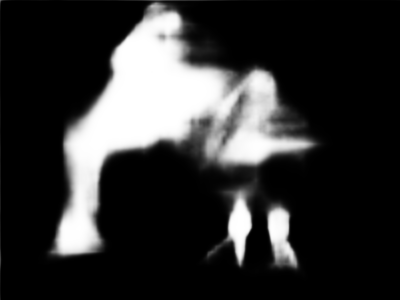}\\
                   \rotatebox{90}{DeepMC} & \hspace{-0.4cm} \includegraphics[height=0.08\linewidth]{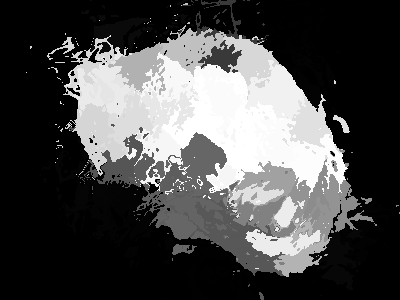}  &\hspace{-0.4cm} \includegraphics[height=0.080 \linewidth]{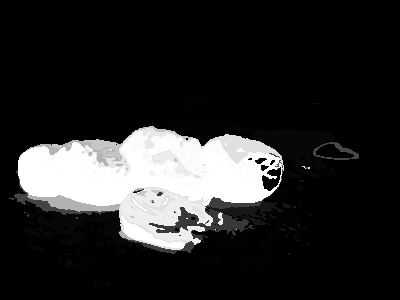}&\hspace{-0.4cm} \includegraphics[height=0.080\linewidth]{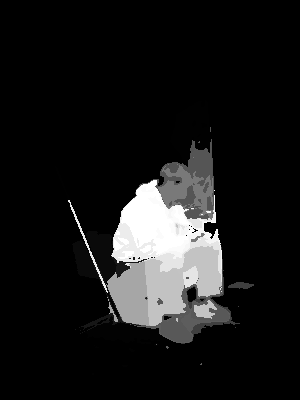} & \hspace{-0.38cm} \includegraphics[height=0.080\linewidth]
{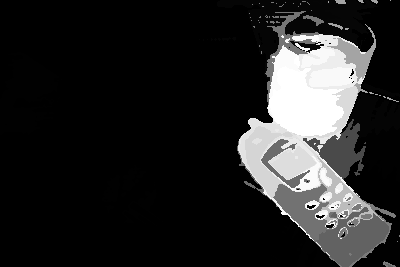}&\hspace{-0.28cm}\includegraphics[height=0.080\linewidth]{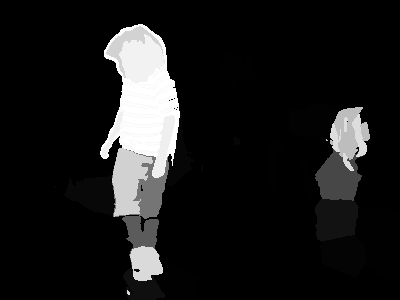}&\hspace{-0.38cm}  \includegraphics[height=0.080\linewidth]{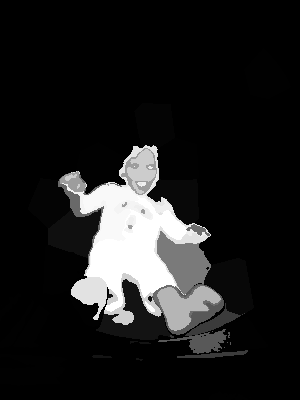}
  &\hspace{-0.4cm} \includegraphics[height=0.080\linewidth]{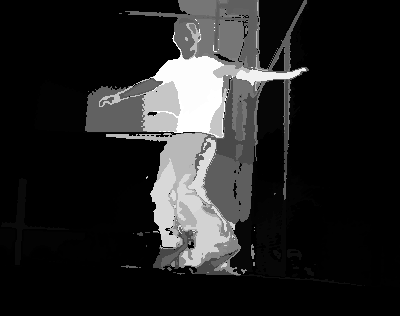}
  &\hspace{-0.4cm}    \includegraphics[height=0.080\linewidth]{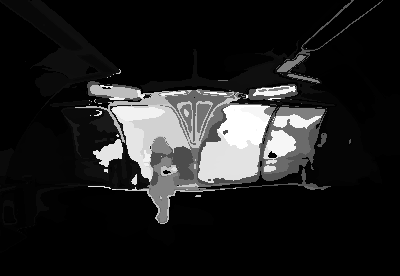}
   & \hspace{-0.4cm}    \includegraphics[height=0.080\linewidth]{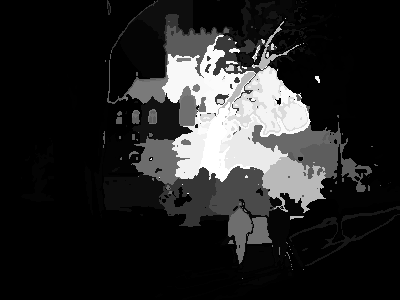}\\
                     \rotatebox{90}{DRFI} & \hspace{-0.4cm} \includegraphics[height=0.08\linewidth]{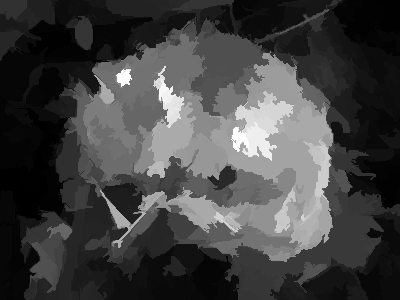}  &\hspace{-0.4cm} \includegraphics[height=0.080 \linewidth]{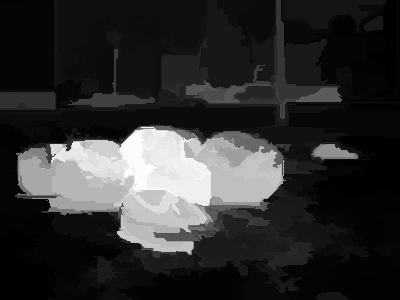}&\hspace{-0.4cm} \includegraphics[height=0.080\linewidth]{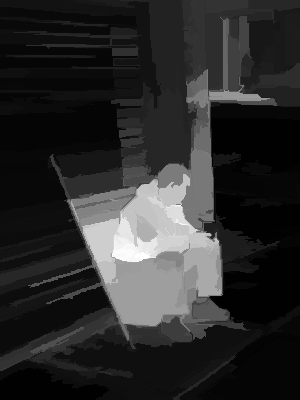} & \hspace{-0.38cm} \includegraphics[height=0.080\linewidth]
{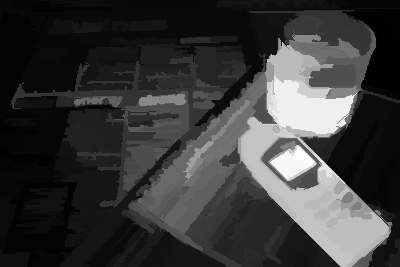}&\hspace{-0.28cm}\includegraphics[height=0.080\linewidth]{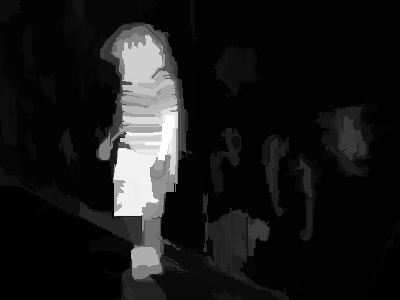}&\hspace{-0.38cm}  \includegraphics[height=0.080\linewidth]{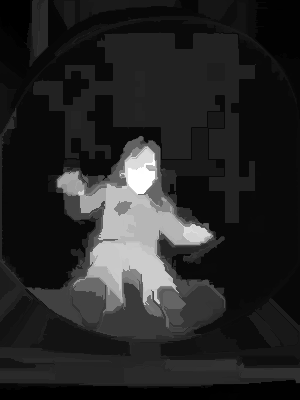}
  &\hspace{-0.4cm} \includegraphics[height=0.080\linewidth]{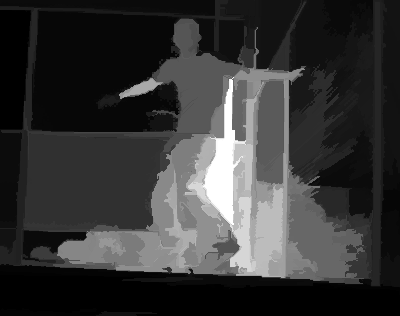}
  &\hspace{-0.4cm}    \includegraphics[height=0.080\linewidth]{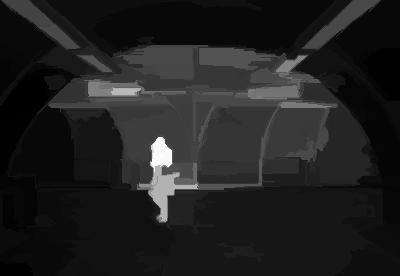}
   & \hspace{-0.4cm}    \includegraphics[height=0.080\linewidth]{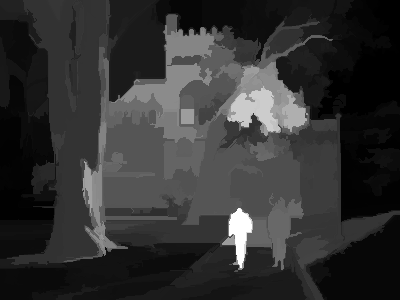}\\
                        \rotatebox{90}{RBD} & \hspace{-0.4cm} \includegraphics[height=0.08\linewidth]{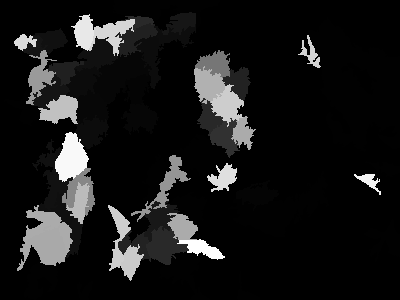}  &\hspace{-0.4cm} \includegraphics[height=0.080 \linewidth]{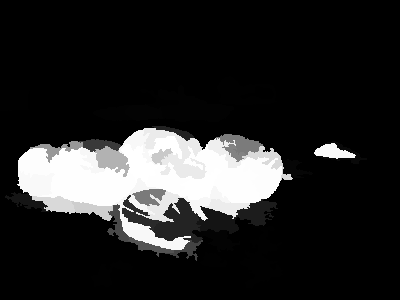}&\hspace{-0.4cm} \includegraphics[height=0.080\linewidth]{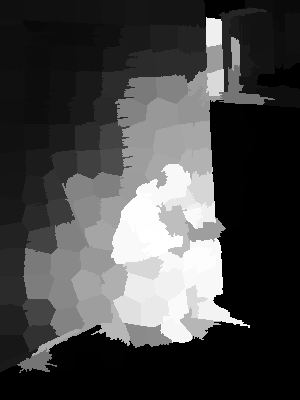} & \hspace{-0.38cm} \includegraphics[height=0.080\linewidth]
{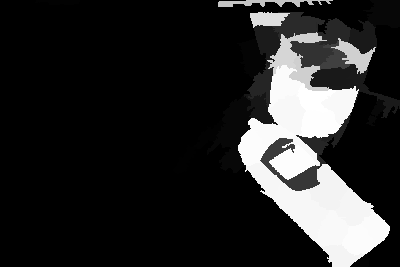}&\hspace{-0.28cm}\includegraphics[height=0.080\linewidth]{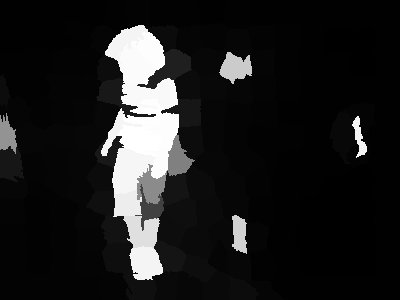}&\hspace{-0.38cm}  \includegraphics[height=0.080\linewidth]{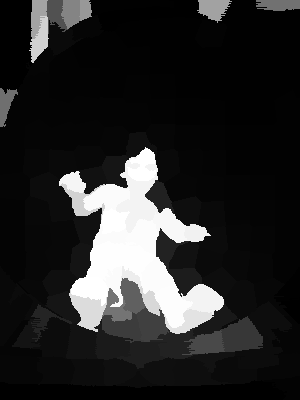}
  &\hspace{-0.4cm} \includegraphics[height=0.080\linewidth]{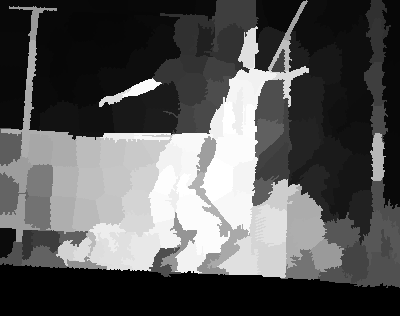}
  &\hspace{-0.4cm}    \includegraphics[height=0.080\linewidth]{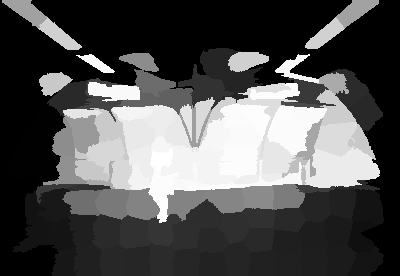}
   & \hspace{-0.4cm}    \includegraphics[height=0.080\linewidth]{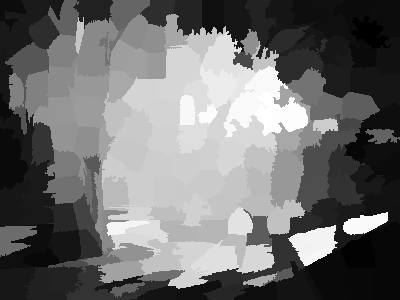}\\
\end{tabular}
\caption{Visual comparison of our method with the other comparing methods. From top to bottom: Input image, ground truth, result of our method, DISC\cite{DISC}, LEGS\cite{LEGS}, MDF\cite{MDF:CVPR-2015}, RFCN\cite{RFCN}, DMT\cite{TIP}, DC\cite{DC}, DeepMC \cite{DeepMC}, DSS\cite{ChengCVPR17}, DRFI\cite{DRFI:CVPR-2013} and RBD\cite{Background-Detection:CVPR-2014}. }
\label{fig:sample_show}
\end{center}
\end{figure*}

\textbf{Qualitative Comparisons:} Figure~\ref{fig:sample_show} demonstrates several visual comparisons, where our method consistently outperforms the competing methods. The first image is in very low contrast, where most of the competing methods failed to capture the salient object, especially for RBD \cite{Background-Detection:CVPR-2014}, while our method captures the salient objects with sharper edge preserved. The 5th image is a simple scenario, and most of the competing methods can achieve good results, while our method achieves the best result with most of the background region suppressed. The 7th image has strong inter-contrast, which leads to quite false detection especially for LEGS \cite{LEGS}, MDF \cite{MDF:CVPR-2015} and RBD \cite{Background-Detection:CVPR-2014}, and our method achieves consistently better results inside the salient object. The salient objects in the last two images are quite small, and the competing methods failed to capture salient regions, while our method capture the whole salient region with high precision.

\begin{figure*}[!htp]   \renewcommand{\arraystretch}{-0.8}
   \begin{center}
   \vspace{-2mm}
   \begin{tabular} {c@{ }  c@{ } c@{ } c@{ }}
   {\includegraphics[width=0.245\linewidth]{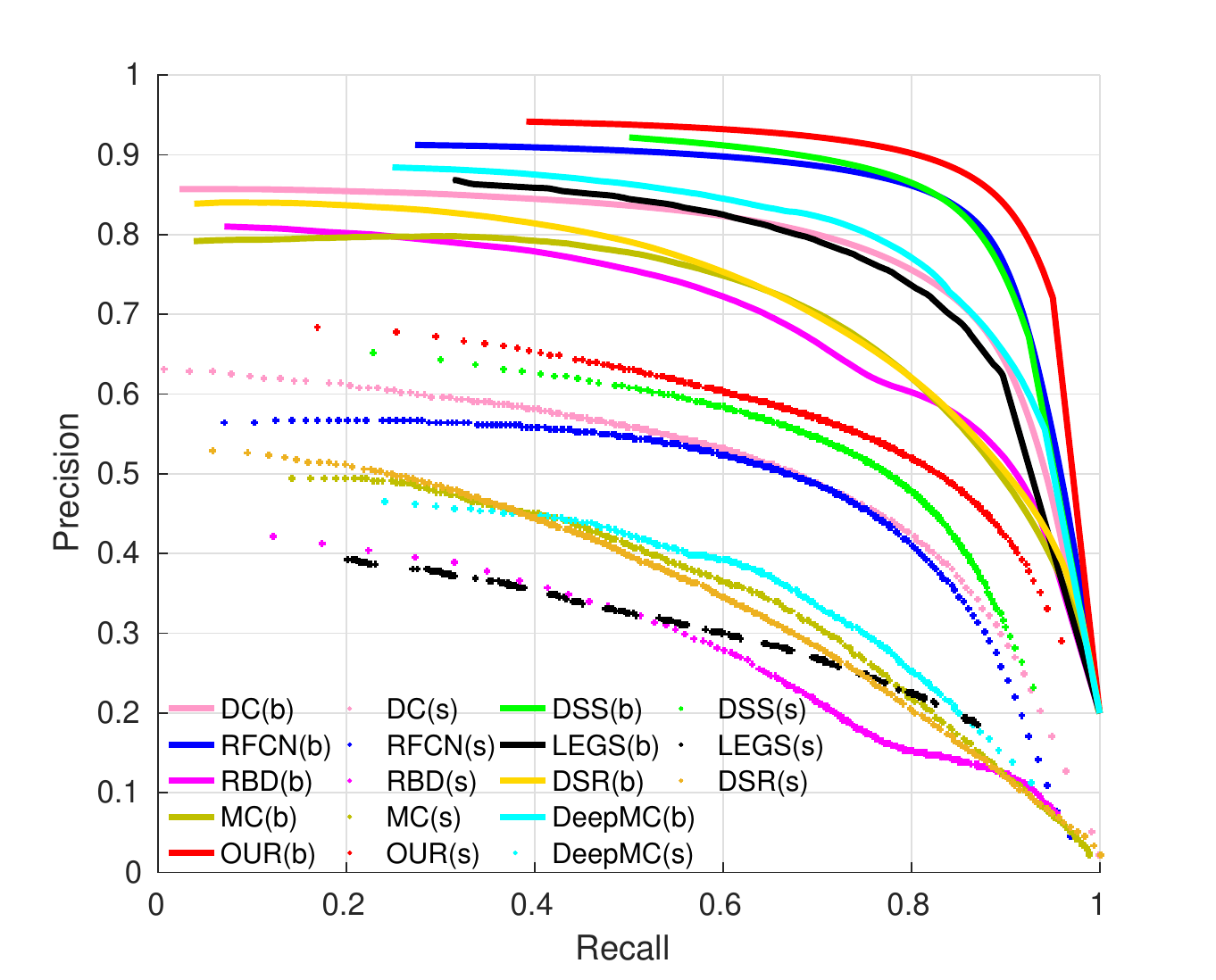}}
      &{\includegraphics[width=0.245\linewidth]{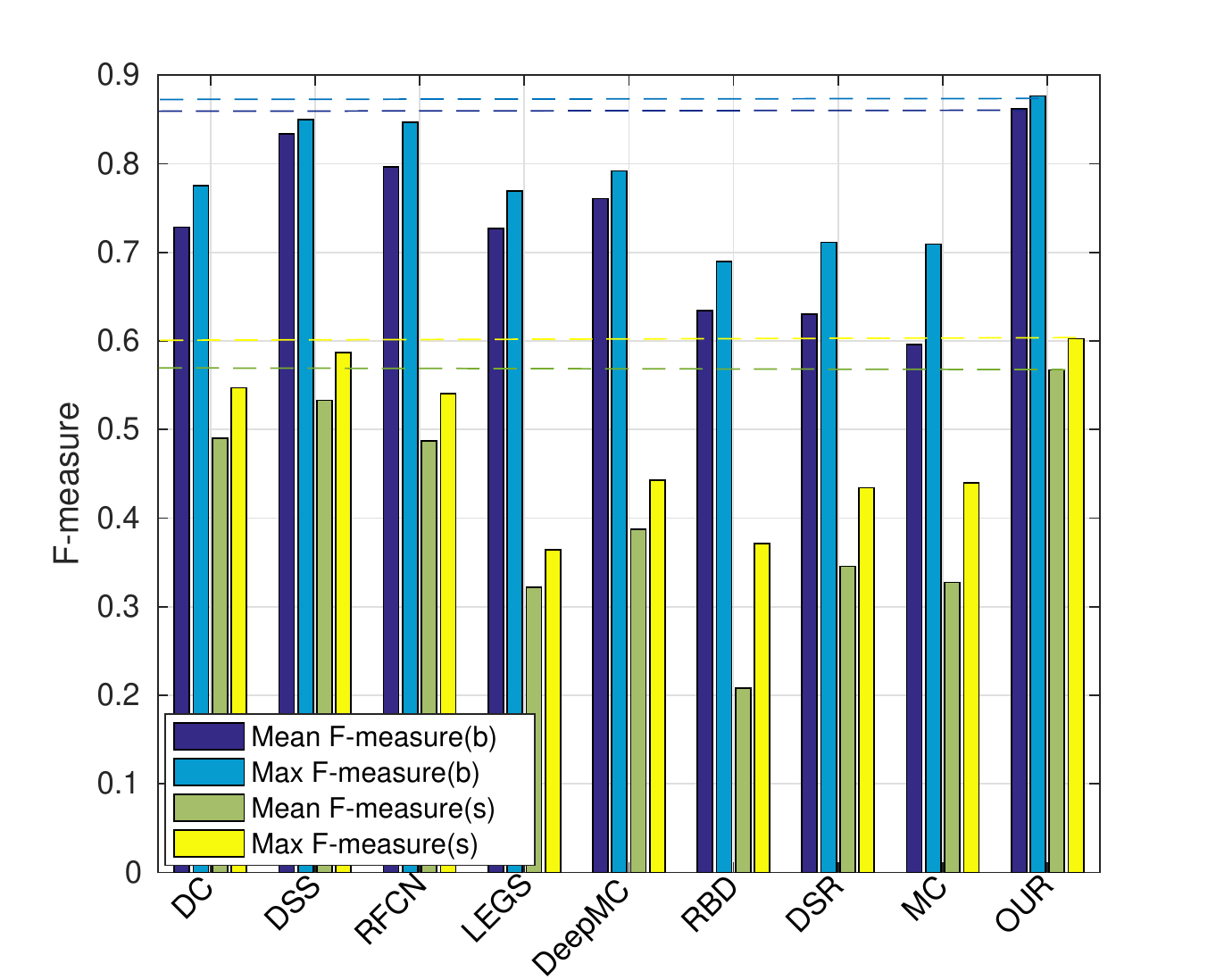}}
            &{\includegraphics[width=0.245\linewidth]{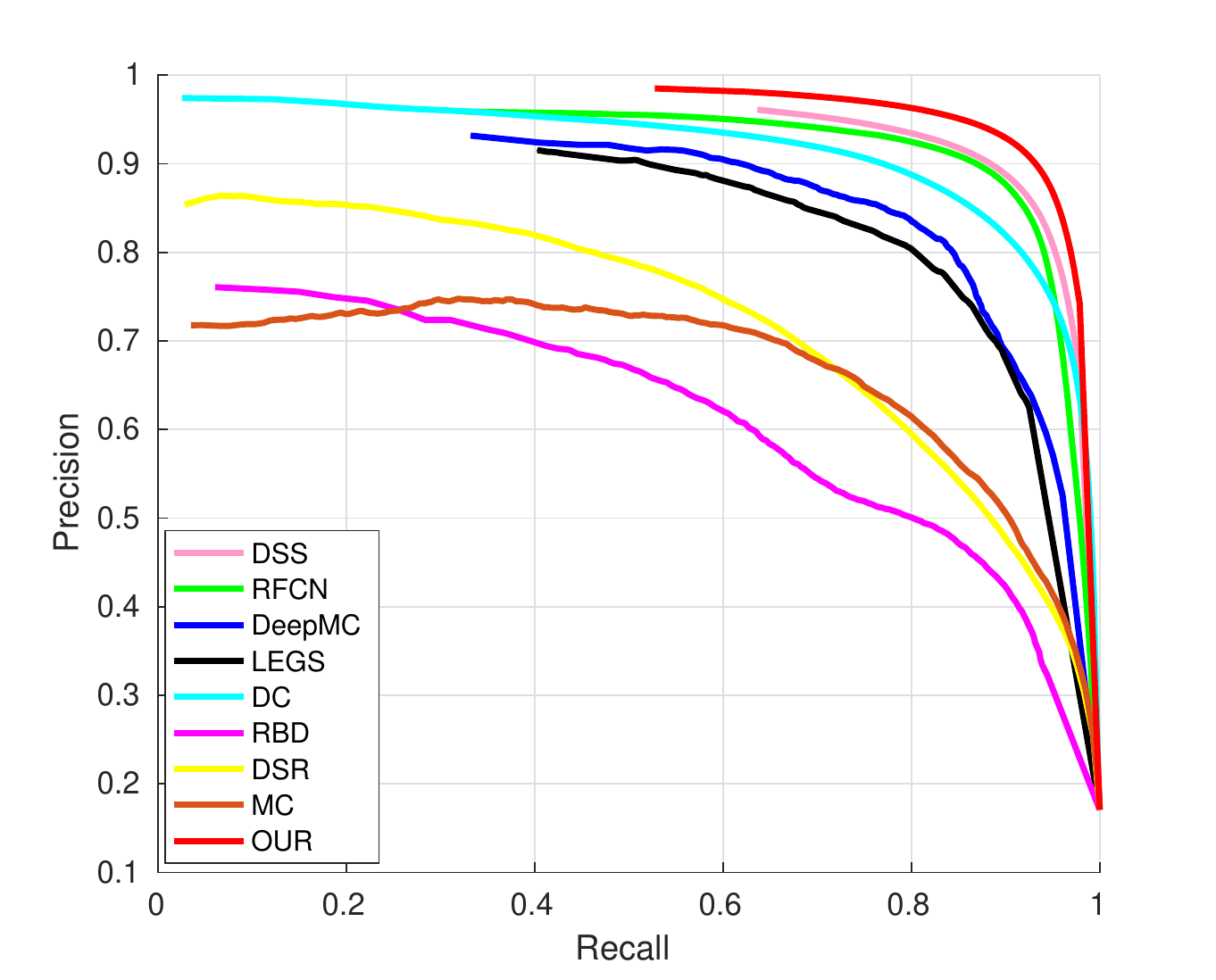}}
         &{\includegraphics[width=0.245\linewidth]{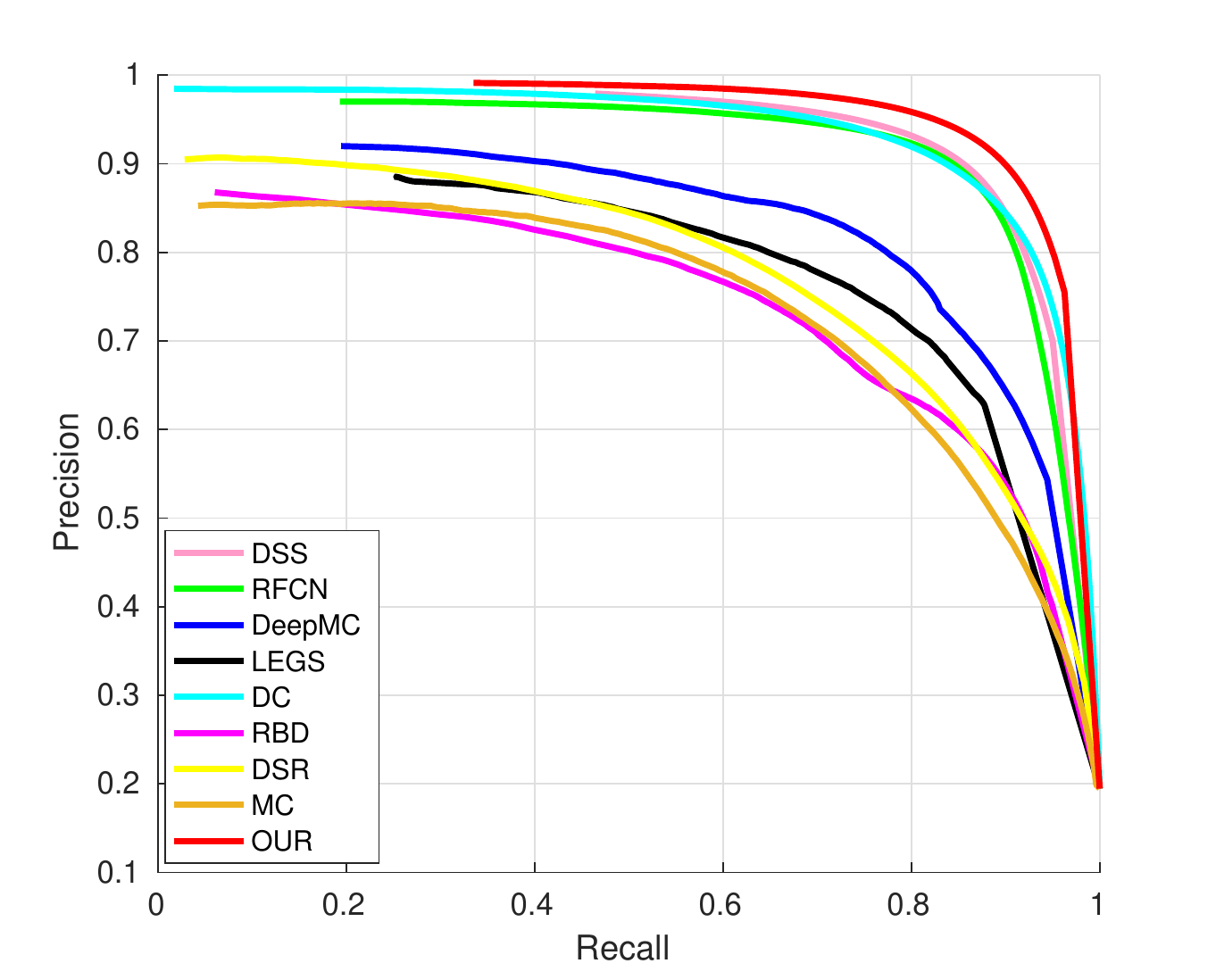}} \\
         \vspace{2mm}
         \small{(a)} & \small{(b)} & \small{(c)} & \small{(d)}
   \end{tabular}
   \end{center}
   \vspace{-2mm}
   \caption{Performance comparison for ablation studies. (a) and (b) are PR curves and F-measure of competing methods on redivided big and small salient object dataset respectively. (c) and (d) are PR curves of competing methods on single and multiple salient object dataset respectively. For both configurations, our method consistently outperforms state-of-the-art saliency detection methods. Best Viewed on Screen.}
   \label{fig:simple_vs_complex}
\end{figure*}

\subsection{Ablation Study}
\label{subsec:ablation_analysis}
To analyze the role of different components in our model, we performed the following ablation studies.

\emph{1) Different training datasets:}
To validate how different training datasets can affect the performance, we trained two extra models with different training datasets. For the first one, we chose MSRA10K \cite{ChengPAMI15} as the training dataset, namely ``MSRA10K'', which contains 10,000 images. We trained this model to analyze whether more training images can lead to better performance. For the second one, we used the HKU-IS dataset \cite{MDF:CVPR-2015} to train our second model, namely ``HKU-IS'', to verify whether model trained on complex training dataset can generate well to other scenarios. Results are shown in Table~\ref{tab:training_dataset_Performance_Comparison}.

\begin{table*}[!htp] \small
\begin{center}
\vspace{-2mm}
\caption{Performance comparisons with different training datasets. Each cell: max F-measure (higher better) / mean F-measure (higher better) / MAE (lower better). (Best in bold)} \centering
\begin{tabular}{l| c| c| c| c| c| c| c| c| c| c}\hline
  & MSRA-B  & ECSSD & DUT & SED1 & SED2 & PASCAL-S & ICOSEG & HKU-IS & THUR & SOD \\ \hline
  &\textbf{0.9310} & 0.9233 & \textbf{0.8010}& 0.9237 & 0.8798& 0.8671 & 0.8617 & \textbf{0.9187}& 0.7811& \textbf{0.8562}\\
Ours & \textbf{0.9184} &0.9082 & \textbf{0.7829}& 0.9085 & 0.8519& 0.8504 & 0.8444 & \textbf{0.9002}& \textbf{0.7638}& \textbf{0.8367}\\
&\textbf{0.0379} & 0.0619 & \textbf{0.0600}& \textbf{0.0627}& 0.0861& 0.1452 & 0.0751 & \textbf{0.0466}& \textbf{0.0673}& 0.1014\\ \hline
  &-  & 0.9142 & 0.7937 & \textbf{0.9283} & 0.8736& 0.8464 & 0.8562 & 0.9111&0.7756 & 0.8356\\
 MSRA10K & -& 0.8929 & 0.7669& 0.9103 & 0.8478& 0.8217 & 0.8344 & 0.8860& 0.7524& 0.8110\\
 &- & 0.0766 & 0.0649& 0.0636 & 0.0889& 0.1714 & 0.0845 & 0.0545& 0.0716& 0.1253\\ \hline
 &0.9177 & \textbf{0.9237} & 0.7860 & 0.9232 & \textbf{0.8931}& \textbf{0.8850} & \textbf{0.8696} & -&\textbf{0.7821} & 0.8501\\
HKU-IS & 0.9052&  \textbf{0.9107} & 0.7682& \textbf{0.9108} & \textbf{0.8760}& \textbf{0.8525} & \textbf{0.8529} & -& 0.7634& 0.8336\\
&0.0433 & \textbf{0.0516} & 0.0667& 0.0637 & \textbf{0.0719}& \textbf{0.1168} & \textbf{0.0548} & -& 0.0723& \textbf{0.0926}\\ \hline

\end{tabular}
\label{tab:training_dataset_Performance_Comparison}
\end{center}
\end{table*}

From Table~\ref{tab:training_dataset_Performance_Comparison}, we observe that our model trained with 2,500 images outperforms the model trained with the entire MSRA10K dataset \cite{ChengPAMI15}. For the PASCAL-S \cite{PASCALS} and SOD \cite{DRFI:CVPR-2013} datasets, our model achieves more than 2\% improvement in mean F-measure and max F-measure, as well as more than 2\% reduction in MAE compared with the ``MSRA10K'' model, which proves that our training dataset of 2,500 training images from MSRA-B is comprehensive enough for training saliency model. Furthermore, compared with ``HKU-IS'', our model gets better results on relatively simple datasets, and worse result when dataset become complex, which encourages us to train model with training images of more complex scenarios.

\emph{2) Salient objects with diverse scales:} According to Table~\ref{tab:deep_unsuper_Performance_Comparison}, both deep saliency detection methods and handcrafted saliency methods achieve better performance for relatively simple testing dataset (MSRA-B dataset for example, where more than half of images with salient region occupying more than $20\%$ of images), and worse performance for dataset with multiple small salient objects (DUT dataset for example, where more than half of images with salient region occupying less than $10\%$ of images), which illustrates that small salient objects detection is still challenging for those deep learning based methods.

To illustrate that our method trained on simple scenarios can generate well to datasets with multiple small salient objects, we divide the ten testing datasets into big salient objects dataset and small salient objects dataset, where the former one includes 18,455 images, and the later one includes 2,353 images. We define images include less than $1/25$ part of the salient region as small salient objects image. Then we compute mean F-measure, max F-measure and PR curve of each method on this re-divided saliency dataset, and the performance is shown in Fig.~\ref{fig:simple_vs_complex}(a) and (b), where ``Our(b)'' and ``Our(s)'' represent our performance on big and small salient object dataset respectively, ``Mean F-measure(b)'' and ``Mean F-measure(s)'' represent mean F-measure of saliency methods on big and small salient object dataset respectively. We could draw two conclusions from Fig.~\ref{fig:simple_vs_complex}(a) and (b). Firstly, both deep saliency methods and handcrafted saliency methods work better on the big salient objects dataset than on the small salient objects dataset. Secondly, by integrating bottom-layer features and handcrafted feature, our method consistently achieves the best performance for both datasets. The reason for the better performance lies in two parts: 1) bottom sides features in our model have a small receptive field, which works well for small salient objects; 2) our salient edge is scale-aware, which helps to detect small salient objects inside salient edges.

\emph{3) Different numbers of salient objects:} We trained our model on MSRA-B dataset, where most images have a single dominant salient object. To illustrate the generalization ability of our method to multiple salient objects dataset, we divided HKU-IS dataset to two parts (we chose the HKU-IS dataset because most of the images in HKU-IS dataset contain more than one salient objects): single salient object dataset and multiple salient objects dataset, where the former one contains 605 images and the later one contains 3,842 images. We compute the PR curves of our method and the competing methods for both datasets correspondingly, and the performance is shown in Fig.\ref{fig:simple_vs_complex}(c) and (d), where (c) is the performance on the single salient object dataset, and (d) is the performance on multiple salient object dataset. We could draw three conclusions from Fig.\ref{fig:simple_vs_complex}(c) and (d). Firstly, on both situations, our method consistently achieves the best performance which proves the effectiveness of our model. Secondly, compared with single salient object dataset, for those handcrafted saliency methods, they achieve better performance on multiple salient object dataset, and the main reason is due to larger salient region occupation of images in the multiple salient object dataset. Thirdly, the occupation of the salient region remains the key element in achieving better performance for multi-scale saliency detection.

\subsection{Execution Time}

Typically, more accurate results are achieved at the cost of a longer run-time. However, this is not our case, as we achieve the state-of-the-art performance while maintaining efficient runtime. Our method maintains a reasonable runtime of around 0.25 second per image.

\section{Conclusions}
\label{sec:conclusion}
We reformulated saliency detection as a three-category dense labeling problem and introduced an edge-aware model. We demonstrated that with saliency edges as constraints in our formulation, we achieved more accurate saliency map and preserve salient edges. Furthermore, we designed a new deep-shallow fully convolutional neural network based on a novel skip-architecture to integrate both deep and handcrafted features. Our method takes the responses of handcrafted saliency detection and normalized color images as inputs and directly learns a mapping from the inputs to saliency maps. We added a multi-scale context module in our model to further improve edge sharpness and spatial coherence.

Our experimental analysis using 10 benchmark datasets (the largest assessment study reported in the literature) and comparisons to 11 state-of-the-art methods show that our method outperforms all existing approaches with a wide margin.

Results also illustrate that small salient object detection is still a significant challenge. Even though we boost the performance of saliency detection, still more work needs to be done on this front since salient objects are often small in typical pictures. We pursue this direction as our future work.



\bibliographystyle{IEEEtran}
\bibliography{SaliencyBib}

\end{document}